\documentclass[10pt,twocolumn,letterpaper]{article}

\usepackage{arxiv}
\usepackage{times}
\usepackage{epsfig}
\usepackage{graphicx}
\usepackage{grffile}
\usepackage{amsmath}
\usepackage{amssymb}
\usepackage{stfloats}

\usepackage[pagebackref=true,breaklinks=true,letterpaper=true,colorlinks,bookmarks=false]{hyperref}

\arxivfinalcopy 

\def\arxivPaperID{1337} 

\ifarxivfinal\fi
\begin{document}


\newcommand{\mydraft}{false}

\newcommand{\magicfigure}[8]{
{\hfill\hbox{\includegraphics[height=#2,draft=\mydraft]{#4}\hspace{#3}\vbox to #2{\hbox{\includegraphics[width=#1,draft=\mydraft]{#5}\hspace{#3}\includegraphics[width=#1,draft=\mydraft]{#6}}\vfill\hbox{\includegraphics[width=#1,draft=\mydraft]{#7}\hspace{#3}\includegraphics[width=#1,draft=\mydraft]{#8}}}}\hfill}
}

\newcommand{\supermagicfigure}[9]{
{\hfill\hbox{\includegraphics[height=#2,draft=\mydraft]{#3}\hspace{.01in}\vbox to #2{\hbox{\includegraphics[width=#1,draft=\mydraft]{#4}\hspace{.01in}\includegraphics[width=#1,draft=\mydraft]{#5}\hspace{.01in}\includegraphics[width=#1,draft=\mydraft]{#6}}\vfill\hbox{\includegraphics[width=#1,draft=\mydraft]{#7}\hspace{.01in}\includegraphics[width=#1,draft=\mydraft]{#8}\hspace{.01in}\includegraphics[width=#1,draft=\mydraft]{#9}}}}\hfill}
}

\newcommand{\superdupermagicfigure}[9]{
    {\hbox{\hspace{#9}\includegraphics[width=#1,draft=\mydraft]{#3}\hspace{#2}\includegraphics[width=#1,draft=\mydraft]{#4}\hspace{#9}}\vfill
        \hbox{\hspace{#9}\includegraphics[width=#1,draft=\mydraft]{#5}\hspace{#2}\includegraphics[width=#1,draft=\mydraft]{#6}\hspace{#9}}\vfill
        \hbox{\hspace{#9}\includegraphics[width=#1,draft=\mydraft]{#7}\hspace{#2}\includegraphics[width=#1,draft=\mydraft]{#8}\hspace{#9}}\vfill
}
}

\newcommand{\fourfigure}[6]{
{\hfill\hbox{\includegraphics[width=#1,draft=\mydraft]{#3}\hspace{#2}\includegraphics[width=#1,draft=\mydraft]{#4}\hspace{#2}\includegraphics[width=#1,draft=\mydraft]{#5}\hspace{#2}\includegraphics[width=#1,draft=\mydraft]{#6}}\hfill}
}

\newcommand{\fourfigureline}[6]{
{\hbox{\includegraphics[width=#1,draft=\mydraft]{#3}\hspace{#2}\includegraphics[width=#1,draft=\mydraft]{#4}\hspace{#2}\hspace{.1in}\includegraphics[width=#1,draft=\mydraft]{#5}\hspace{#2}\includegraphics[width=#1,draft=\mydraft]{#6}}}
}

\newcommand{\fivefigure}[7]{
{\hfill\hbox{\includegraphics[width=#1,draft=\mydraft]{#3}\hspace{#2}\includegraphics[width=#1,draft=\mydraft]{#4}\hspace{#2}\includegraphics[width=#1,draft=\mydraft]{#5}\hspace{#2}\includegraphics[width=#1,draft=\mydraft]{#6}\hspace{#2}\includegraphics[width=#1,draft=\mydraft]{#7}}\hfill}
}

\newcommand{\sixfigure}[8]{
{\hbox{\includegraphics[width=#1,draft=\mydraft]{#3}\hspace{#2}\includegraphics[width=#1,draft=\mydraft]{#4}\hspace{#2}\includegraphics[width=#1,draft=\mydraft]{#5}\hspace{#2}\includegraphics[width=#1,draft=\mydraft]{#6}\hspace{#2}\includegraphics[width=#1,draft=\mydraft]{#7}\hspace{#2}\includegraphics[width=#1,draft=\mydraft]{#8}}}
}

\newcommand{\threefigure}[5]{
{\hbox{\hspace{.1in}\includegraphics[width=#1,draft=\mydraft]{#3}\hspace{#2}\includegraphics[width=#1,draft=\mydraft]{#4}\hspace{#2}\includegraphics[width=#1,draft=\mydraft]{#5}}}
}

\newcommand{\twofigure}[4]{
{\hbox{\includegraphics[width=#1,draft=\mydraft]{#3}\hspace{#2}\includegraphics[width=#1,draft=\mydraft]{#4}}}
}

\title{Local Geometric Indexing of High Resolution Data for Facial Reconstruction from Sparse Markers}

\author{Matthew Cong\\
Industrial Light \& Magic\\
{\tt\small mcong@ilm.com}
\and
Lana Lan\\
Industrial Light \& Magic\\
{\tt\small llan@ilm.com}
\and
Ronald Fedkiw\\
Stanford University\\
Industrial Light \& Magic\\
{\tt\small rfedkiw@stanford.edu}
}

\maketitle

\begin{abstract}
When considering sparse motion capture marker data, one typically struggles to balance its overfitting via a high dimensional blendshape system versus underfitting caused by smoothness constraints.
With the current trend towards using more and more data, our aim is not to fit the motion capture markers with a parameterized (blendshape) model or to smoothly interpolate a surface through the marker positions, but rather to find an instance in the high resolution dataset that contains local geometry to fit each marker.
Just as is true for typical machine learning applications, this approach benefits from a plethora of data, and thus we also consider augmenting the dataset via specially designed physical simulations that target the high resolution dataset such that the simulation output lies on the same so-called manifold as the data targeted.

\end{abstract}

\section{Introduction}

Realistic facial animation has a wide variety of applications in both computer vision and the entertainment industry \cite{williams2006performance}.
It is typically achieved through a combination of keyframe animation, where an animator hand-adjusts controls corresponding to the motion of different parts of the face, and facial performance capture, which uses computer vision to track the motion of an actor's face recorded from one or more cameras.
Despite the many techniques developed over the years, facial performance capture remains a difficult task, and the high degree of accuracy required to generate realistic facial animation severely suppresses its widespread impact.

One class of techniques which has a proven track record uses markers painted on an actor's face in conjunction with a stereo pair of head mounted cameras \cite{bhat2013high, williams2006performance}.
These markers are tracked in each camera and triangulated to obtain a sparse set of animated 3D bundle positions representing the motion of the actor's face.  
In order to reconstruct a full 3D facial pose for each frame of 3D bundles, one often uses a parameterized (blendshape) model \cite{lewis2014practice, parke1974parametric}.
However, these parameterized models often have large infeasible spaces.
While a skilled animator can aim to avoid these infeasible spaces, an optimization algorithm would need them explicitly specified which is typically not practical.
Another commonly used approach interpolates bundle displacements across the face \cite{bhat2013high}.
However, this results in reconstructed geometry that is overly smooth since the sparse bundle positions cannot represent high-resolution details between the bundles, especially details that appear during expressions, e.g.~folds, furrows, and wrinkles.
To address these shortcomings, we follow the current trend in the deep learning community of adding more and more data by using a large dataset of facial shapes to inform the reconstruction of the face surface geometry from the tracked bundle positions.

Our approach to this problem can be thought of as local geometric indexing wherein each bundle needs to identify relevant associated geometry from the dataset.
To accomplish this, we envision the dataset as a separate point cloud for each bundle; this point cloud is obtained by evaluating the 3D position of the relevant bundle on each face shape in the dataset.
These point clouds are then used to index the dataset in order to figure out the most relevant shapes given a bundle position.
A bundle position that lies outside of its associated point cloud indicates a lack of data and can be projected back towards the point cloud.
On the other hand, it is also possible for many candidate points to exist in the point cloud in which case neighboring bundles and their associated point clouds can be used to disambiguate.
Finally, the shapes chosen for each bundle are combined to obtain a high-resolution dense reconstruction of the facial geometry.

We begin the exposition by describing the creation of our facial shape dataset, which is initially bootstrapped via a combination of dense performance capture and hand sculpting for a small set of expressions and is further augmented using physical simulation.
Then, we detail our local geometric indexing scheme and show how it can be used to find the shapes that are most relevant to a bundle given its position.
This is followed by a discussion of the various smoothness considerations that are used to inform our approach for spatially blending the relevant shapes across the face to recover a high-resolution dense reconstruction of the full face.
Finally, we apply our algorithm to a series of feature film production examples and compare the results to other popular approaches.






\section{Prior Work} \label{sec:prior_work}

\paragraph{Capture:}
High-resolution facial geometry can be captured using dense performance capture techniques such as \cite{beeler2010high, beeler2011high, fyffe2014driving, ghosh2011multiview}.
However, these methods typically require environments with controlled lighting and dedicated camera hardware.
These restrictions, along with the limitations on the actor's motion, often make these techniques unsuitable for on-set capture where an actor often needs to interact with the set and/or other actors.
On-set capture typically involves painting a marker pattern on an actor's face and recording the actor's performance with a set of helmet mounted cameras.
The markers can be tracked in the resulting footage and triangulated to recover a sparse set of bundle positions that follow the actor's facial performance.

\vspace{-0.1in}
\paragraph{Reconstruction:}
In order to animate the neutral mesh of an actor, one could compute a smooth deformation of the face mesh by interpolating the motion of the bundles (see e.g.~the corrective shape computed in \cite{bhat2013high} and the non-rigid ICP approach of \cite{li2009robust}).
However, this usually results in a deformed mesh that contains too much of the high-frequency detail of the neutral mesh and too little of the high-frequency detail associated with a particular expression.
In order to add high frequency details to the reuslting deformed mesh, \cite{li2009robust} projects the smoothly deformed mesh towards per-frame dense scans.
Several approaches that do not require additional per-frame scans have also been proposed including \cite{bermano2014facial}, \cite{bickel2007multiscale}, \cite{bickel2008pose}, \cite{ma2008facial}, and \cite{huynh2018mesoscopic} which use deformation gradients, a nonlinear shell energy, feature graph edge strains, polynomial displacement maps, and neural networks respectively.
Masquerade \cite{moser2017masquerade} combines some of these approaches for facial performances solved from helmet mounted cameras.
However, such approaches may not remove high-frequency details in the neutral mesh that are not present in the expression.
Furthermore, if the smooth deformation interpolates the bundles, the addition of fine scale details in this manner can potentially move the surface farther away from the bundles.

\vspace{-0.1in}
\paragraph{Blendshapes:}
Instead of interpolating the motion of the bundles directly, one could use the markers and/or bundles to drive a blendshape facial rig \cite{lewis2014practice} which specifies the deformation of the face as a linear combination of facial shapes.
These facial shapes are acquired using dense performance capture (see e.g.~\cite{beeler2010high, beeler2011high, fyffe2014driving, ghosh2011multiview}) and/or sculpted by an experienced modeler \cite{cong2017muscle, lan2017lessons}.
Then, one can optimize for the shape weights that minimize the differences between the marker and bundle positions and their associated projected surface positions and surface positions respectively on the resulting mesh \cite{bhat2013high, cao2013shape, cao2014displaced, li2010example}.
Alternatively, one could minimize the difference between synthetic renderings of the face and the corresponding input images (see e.g.~\cite{thies2018facevr, kim2018deep}).
However, such approaches often result in unnatural combinations of shapes with weights that are difficult to interpret \cite{bouaziz2013online, chuang2002performance, ribera2017facial, seol2012spacetime}.
These infeasible combinations can be avoided by experienced animators but are extremely problematic for optimization algorithms.
In order for an optimization algorithm to avoid these combinations, one would need to specify all such invalid combinations in the high-dimensional Cartesian space of facial shapes, which is intractable.


\vspace{-0.1in}
\paragraph{Patch-Based Approaches:}
The patch-based model of \cite{wu2016anatomically} is particularly notable because it uses a smaller number of facial shapes compared to a traditional blendshape rig.
Despite the small number of facial shapes, the resulting per-patch shape in this model still lies in the Cartesian product of the input shapes.
Thus, as the size of the dataset increases, one would still expect the model to overfit on a per-patch basis.
The FaceIK editing technique of \cite{zhang2004spacetime} also uses a localized blendshape deformation model by adaptively segmenting the face mesh based on user specified control points, solving for blendshape weights for each control point based on its position, and spatially blending the resulting weights across the mesh using a radial basis function.
In order to improve sparsity of the blendshape weights and reduce overfitting, blendshapes that are farther away from the control points are penalized.
Unlike \cite{zhang2004spacetime}, which uses an interpolatory approach, our approach uses a non-manifold mesh and other considerations to boost the domain from $\mathbb{R}^3$ into higher dimensions.
Other localized models have also been proposed such as \cite{tena2011interactive} which uses PCA-based patches.




\section{Dataset} \label{sec:dataset}

Given the high-resolution mesh of an actor in the neutral or rest pose, we construct a dataset of high-quality facial shapes that sufficiently samples the actor's range of motion and expression.
We bootstrap this process by acquiring high-resolution facial geometry for a selection of the actor's (extreme) facial poses taken from a range of motion exercise using the Medusa performance capture system \cite{beeler2010high, beeler2014rigid, beeler2011high}.
For each facial pose, Medusa both deforms the neutral mesh to the pose based on images from multiple cameras and estimates the cranium rigid frame associated with the deformed mesh.
The cranium rigid frame is manually refined (if necessary), validated against the images from each of the cameras, and then used to stabilize the associated deformed mesh.
Each stabilized deformed mesh is then stored as a per-vertex displacement from the neutral mesh.

These stabilized facial shapes are further improved using physical simulation.
Starting from the high-resolution neutral mesh, we build a simulatable anatomical face model by morphing an anatomically and biomechanically accurate template model following the approach of \cite{cong2015fully}.
Then, we use the art-directed muscle simulation framework of \cite{cong2016art} to target each captured facial shape to obtain a corresponding simulated facial shape with improved volume conservation, more realistic stretching, and a more plausible response to contact and collision.
The captured and simulated facial shape are then selectively blended together by a modeler to obtain a combined facial shape that incorporates both the high degree of detail obtained from capture as well as the physical accuracy obtained from simulation.
Finally, this combined facial shape is further refined by a modeler based on the images in order to resolve any remaining artifacts before being added to the dataset.
See \cite{cong2017muscle, lan2017lessons}.

At this point, the dataset consists of facial shapes corresponding to various extreme poses.
We augment the dataset with in-betweens to better represent subtle motions and combinations of expressions.
To do this, one could construct a blendshape system using the facial shapes already in the dataset and evaluate this blendshape system at fixed intervals in the high-dimensional Cartesian space; however, the resulting in-betweens would suffer from well-known linear blendshape artifacts such as volume loss.
Instead, one could use the aforementioned process targeting the high-dimensional Cartesian space blendshapes with the art-directed muscle simulation framework of \cite{cong2016art}, or alternatively one could use the approach of \cite{cong2016art} alone to move between various extreme facial poses creating in-betweens.
We utilize a combination of these options to add anatomically motivated nonlinear in-betweens to the dataset.



\section{Local Geometric Indexing}

Our local geometric indexing scheme begins by constructing a separate point cloud for each bundle, accomplished by evaluating the surface position of the bundle on each facial shape in the dataset.
The brute force version of our algorithm would tetrahedralize each point cloud with all possible combinations of four points resulting in a non-manifold tetrahedralized volume (See Sec.~\ref{subsec:tetrahedralization}).
Then, given a bundle position, we find all the tetrahedra in the associated tetrahedralized volume that contain it.
Since the tetrahedralized volumes are only dependent on the dataset, this process can be accelerated by precomputing a uniform grid spatial acceleration structure \cite{fujimoto1988accelerated, pharr2010physically}.
For each of these tetrahedra, we compute the convex barycentric weights $(w_i, w_j, w_k, w_l)$ of the bundle position and use these to blend together the four facial shapes $\vec{b}_i$, $\vec{b}_j$, $\vec{b}_k$, and $\vec{b}_l$ corresponding to the vertices of the tetrahedron.
The resulting candidate shape is given by
\begin{align}
    \vec{x} = \vec{x}_0 + \sum_{n \in \{i,j,k,l\}} w_n\vec{b}_n \label{eq:piecewise_linear}
\end{align}
where $\vec{x}_0$ represents the neutral mesh positions.
By construction, the candidate surface geometry is guaranteed to intersect the bundle position and lie within the convex hull of the facial shapes.

If there are no tetrahedra that contain the bundle position, we project the bundle position to the convex hull of the associated point cloud by using the barycentric coordinates for the closest point on the associated tetrahedralized volume.
The lack of tetrahedra containing a bundle position indicates a need for additional facial shapes in the dataset; however, this projection approach gives reasonable results in such scenarios.

Local geometric indexing can be viewed as a piecewise linear blendshape system, although the pieces are difficult to describe due to overlapping non-manifold tetrahedra and various nonlinear, nonlocal, and higher dimensional strategies for choosing between multiple overlapping tetrahedra.
Still, by augmenting the dataset with more in-betweens, we can insert so-called Steiner points \cite{berg2008computational} allowing for increased efficacy -- stressing the importance of collecting more and more data.

\subsection{Tetrahedralization} \label{subsec:tetrahedralization}

As the size of the point cloud increases, the construction of all possible tetrahedra quickly becomes unwieldy.
Thus, we aggressively prune redundancies from the point cloud, e.g.~removing points corresponding to expressions that do not involve them.
For example, we do not add bundle evaluations to a forehead bundle's point cloud from expressions that only involve the lower half of the face.
Besides reducing the number of points, we may also eliminate tetrahedra especially those that are poorly shaped: too thin, too much spatial extent, etc.
Moreover, tetrahedra which are known to be problematic admitting shapes that are locally off-model can also be deleted.
Similar statements hold for unused or rarely used tetrahedra, etc.
Importantly, through continued use and statistical analysis, our tetrahedral database can evolve for increased efficiency and quality.

Instead of considering all possible combinations of four points, one could tetrahedralize each point cloud using a space-filling tetrahedralization algorithm such as constrained Delaunay tetrahedralization \cite{shewchuk2002constrained}.
However, this would restrict a bundle position to lie uniquely within a single tetrahedron and create a bijection between a bundle position and local surface geometry.
This is problematic because different expressions can map to the same bundle position with different local curvature.
For example, a bundle along the midline of the face on the red lip margin can have the same position during both a smile and a frown.
Thus, it is better to construct an overlapping non-manifold tetrahedralization in order to allow for multiple candidate local surface geometries for a bundle position, later disambiguating using additional criteria.
Moreover, as discussed later, one may create more than one point cloud for an associated bundle with each point cloud corresponding to different criteria.
For example, the shapes one uses for an open jaw could differ significantly when comparing a yawn and an angry yell; different point clouds for sleepy, angry, happy, etc. would help to differentiate in such scenarios.

Again, we stress that a space-filling manifold tetrahedralized volume allows a bundle only three degrees of freedom as it moves through the manifold tetrahedralized volume in $\mathbb{R}^3$, whereas overlapping non-manifold tetrahedra remove uniqueness in $\mathbb{R}^3$ boosting the domain to a higher dimensional space; then, other considerations may be used to ascertain information about other dimensions and select the appropriate tetrahedron.



\section{Smoothness Considerations}

\begin{figure}
    \centering
    \twofigure{.495\linewidth}{.03in}{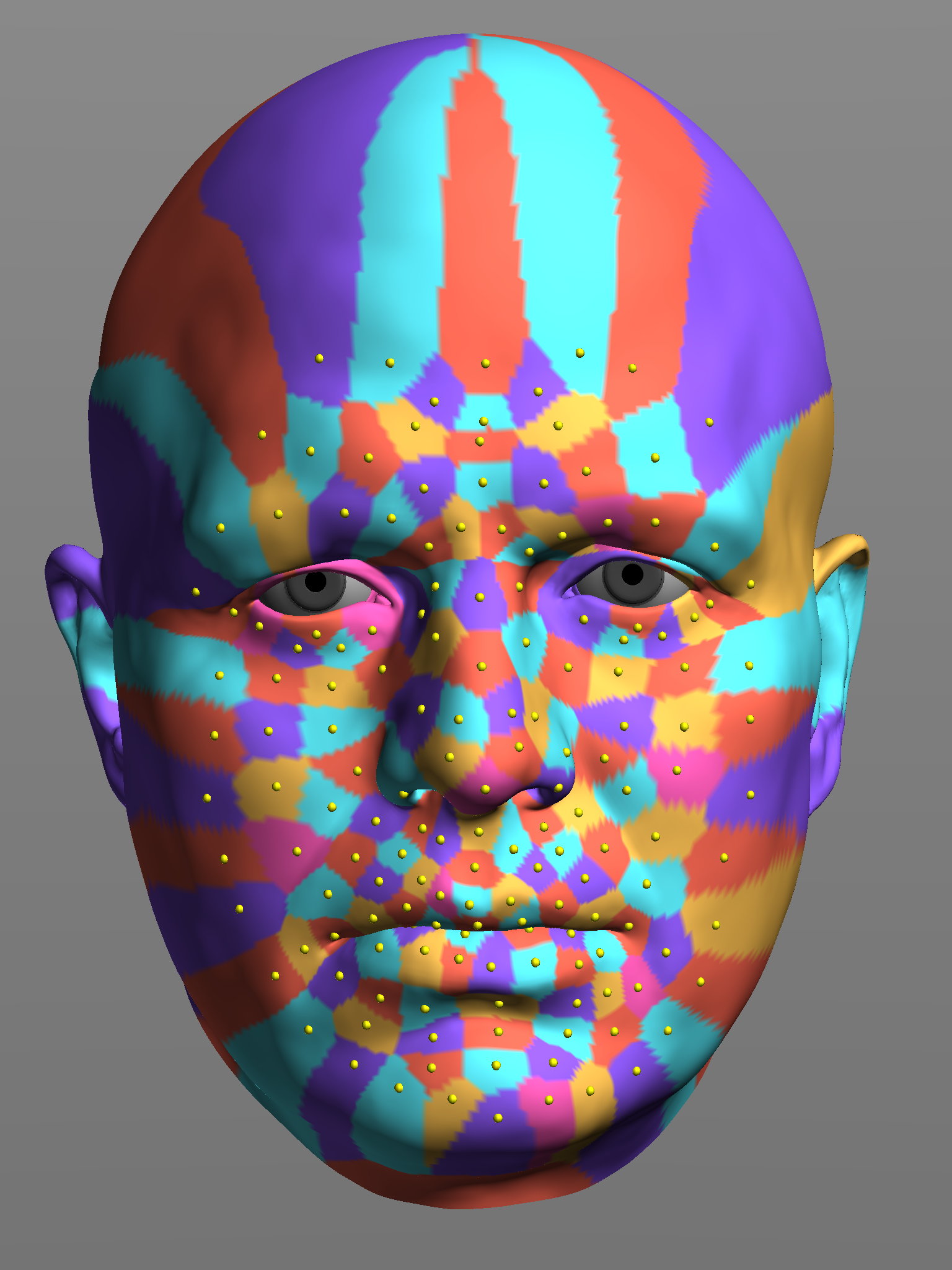}{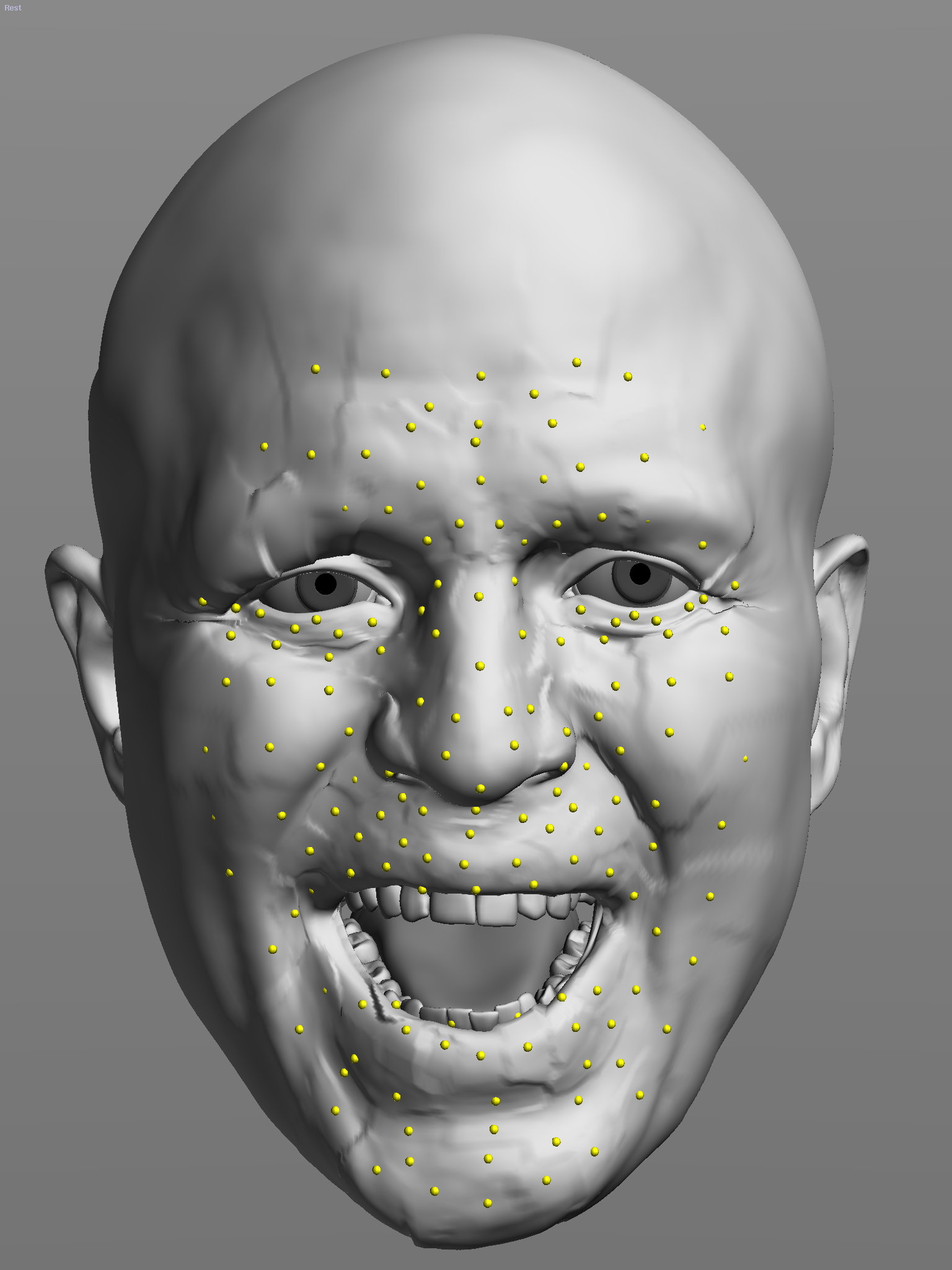}
    \vspace{.02in}
    \twofigure{.495\linewidth}{.03in}{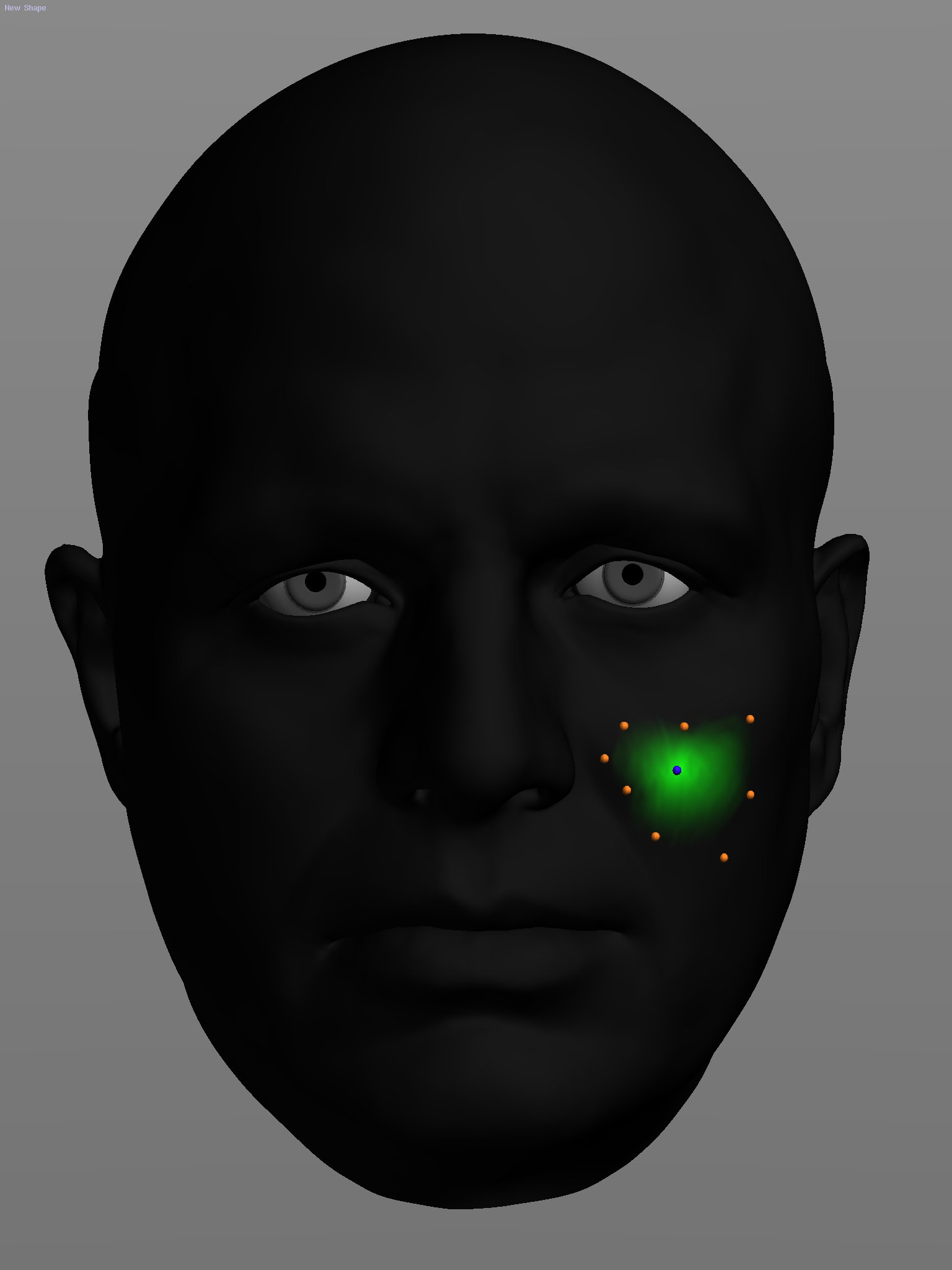}{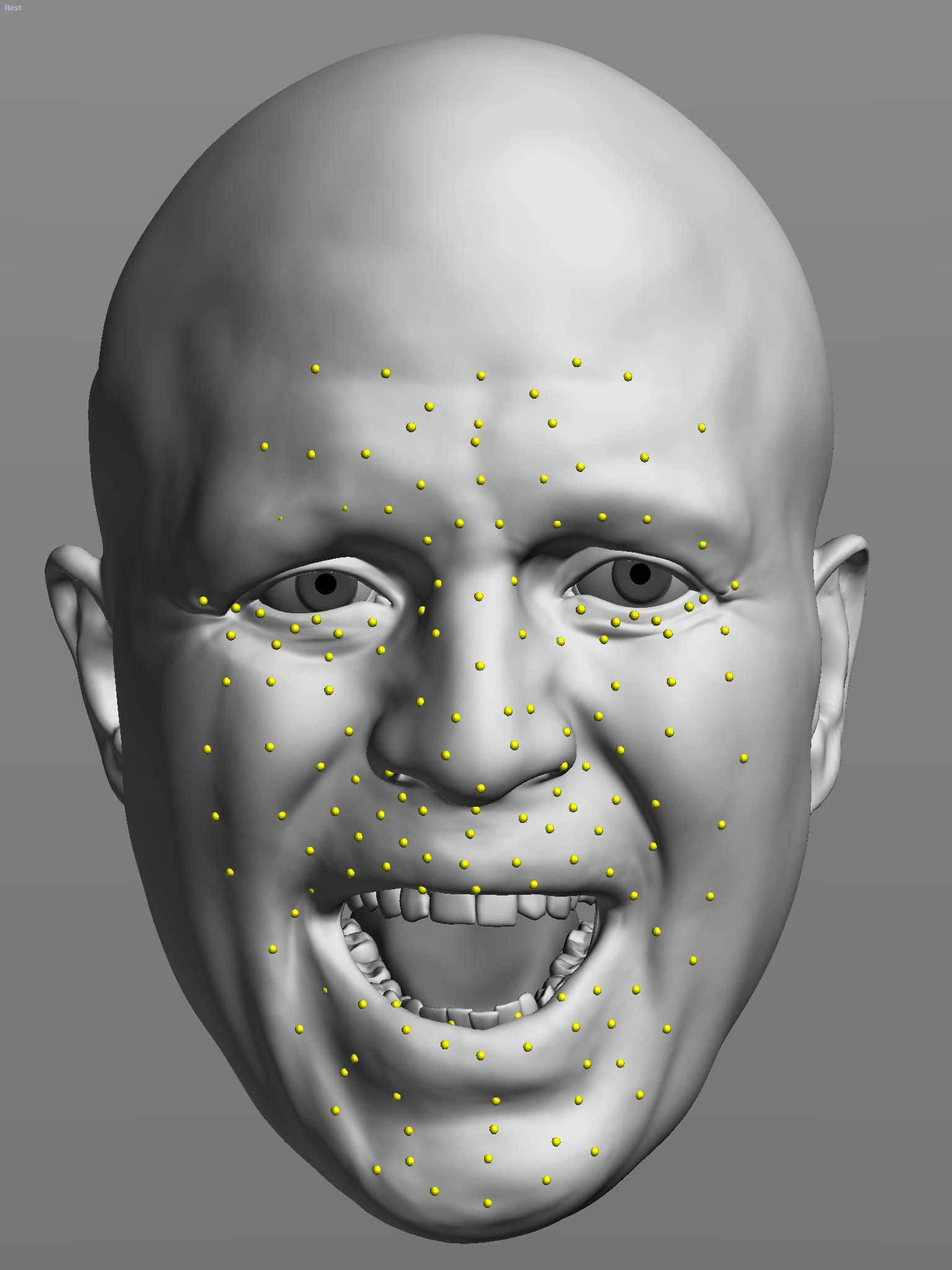}
    \caption{Top Left: Voronoi diagram on the neutral mesh. Top Right: Applying the locally indexed surface geometry to each Voronoi cell results in discontinuities across cell boundaries. Bottom Left: Natural neighbor weights for a single bundle. The weight is 1 at the bundle surface position and 0 at the surface positions corresponding to neighboring bundles. Bottom Right: Using natural neighbor weights, we obtain a smoother reconstruction that interpolates the bundle positions. \label{fig:voronoi_diagram}}
    \vspace{-0.2in}
\end{figure}

Our local geometric indexing scheme generates local surface geometry for each bundle independently, and we subsequently sew the local surface geometry together to create a unified reconstruction of the full face.
Because only local geometry is required, we only need to store small surface patches (and not the full face geometry) for each point in the point cloud making the method more scalable.
To sew the local patches together, we first construct a Voronoi diagram on the neutral mesh using the geodesic distances to the surface position of each bundle in the rest pose. See Figure~\ref{fig:voronoi_diagram} (Top Left).
These geodesic distances are computed using the fast marching method \cite{kimmel1998computing}. 
The local surface geometry for each bundle could then be applied to its associated Voronoi cell on the mesh, although the resulting face shape would typically have discontinuities across Voronoi cell boundaries as shown in Figure~\ref{fig:voronoi_diagram} (Top Right).

We have experimented with a number of scattered interpolation methods aimed at smoothing the local patches across Voronoi cell faces including, for example, radial basis functions as in \cite{wu2016anatomically}.
We experimentally achieved the best results leveraging our Voronoi diagram using natural neighbor interpolation \cite{park2006discrete, sibson1980vector}.
For a given vertex on the neutral mesh, natural neighbor weights are computed by inserting the vertex into the precomputed Voronoi diagram, computing the areas stolen by the new vertex's Voronoi cell from each of the pre-existing neighboring Voronoi cells, and normalizing by the total stolen area.
For each vertex, the natural neighbor weights are used to linearly blend the shapes used for each surrounding bundle.
Note that a vertex placed at a bundle position would not change the Voronoi regions of surrounding bundles and would merely adopt the Voronoi region from the bundle it is coincident with; this guarantees that the resulting blended surface still exactly interpolates the bundle positions.
In this way, we obtain a $C^0$ continuous reconstructed surface \cite{piper1993properties} that passes through all of the bundle positions. See Figure~\ref{fig:voronoi_diagram} (Bottom Right).
We found that constructing the Voronoi diagram and calculating the natural neighbor weights in UV/texture space and subsequently mapping them back onto the 3D mesh yielded smoother natural neighbor weights than performing the equivalent operations on the 3D mesh directly.

\subsection{Choosing Tetrahedra} \label{subsec:choosing_tetrahedra}

In order to minimize kinks in the $C^0$ continuous reconstructed surface, we use an additional smoothness criterion when choosing between overlapping tetrahedra.
If there are multiple tetrahedra which contain the bundle position, we choose the tetrahedron that results in local surface geometry that minimizes the distances from neighboring bundle positions to their respective surface positions.
This indicates that the local surface geometry is representative of the bundle as well as the neighborhood between the bundle and its neighboring bundles.

In the case where no tetrahedra contain the bundle position, one can apply a similar criterion to project the bundle back to the dataset in a smooth manner.
When deciding which tetrahedron to project to, one could consider not only the distance from the bundle under consideration to the resulting surface, but also the distances that neighboring bundles would be from the resulting surface.

In the case of an animated bundle with time-varying position, we apply additional criteria to prevent disjoint sets of shapes from being chosen in neighboring frames, ameliorating undesirable oscillations in the animated reconstructed surface.
To do this, we assign higher priority to tetrahedra which share more points and therefore facial shapes with the tetrahedron used on the previous frame, biasing towards a continuous so-called winding number on the non-manifold representation.


\section{Jaw Articulation} \label{sec:jaw_articulation}

So far, we have considered facial shapes and bundle positions relative to the neutral mesh.
However, these shapes and bundle positions may include displacements due to rotational and prismatic jaw motion \cite{sifakis2005automatic, zoss2018empirical}.
This can result in significant linearized rotation artifacts in the reconstruction which reduces the generalizability of our approach.
In order to address this, we hybridize our approach by using linear blend skinning to account for the jaw pose.

To do this, we modify Eq.~\ref{eq:piecewise_linear} with a block diagonal matrix of spatially varying invertible transformations $T(\theta)$ calculated using linear blend skinning from the jaw parameters $\theta$ and a set of unskinned facial shapes $\vec{b}^*_n$ to obtain
\begin{align}
    \vec{x} = T(\theta)\left(\vec{x}_0 + \sum_{n \in \{i,j,k,l\}} w_n\vec{b}^*_n\right). \label{eq:hybrid_piecewise_linear}
\end{align}
For a shape with known jaw parameters $\theta_n$, setting Eq.~\ref{eq:piecewise_linear} equal to Eq.~\ref{eq:hybrid_piecewise_linear} and rearranging terms gives an expression for the unskinned facial shape
\begin{align*}
    \vec{b}^*_n = T(\theta_n)^{-1}\left(\vec{x}_0 + \vec{b}_n\right) - \vec{x}_0
\end{align*}
as a function of the facial shape $\vec{b}_n$. See \cite{lewis2014practice, orvalho2012facial}.
In order to utilize this approach, every shape in the database needs the jaw parameters $\theta_n$ estimated so that we may store $\vec{b}^*_n$ instead of $\vec{b}_n$.
Similarly for each frame, $\theta$ must be estimated using one of the usual methods for head and jaw tracking so that the bundle positions can be unskinned before indexing into the point cloud.

As mentioned in Sec.~\ref{subsec:tetrahedralization}, having a large number of points can result in an unwieldy number of tetrahedra.
Thus, one could bin points into different point clouds based on a partition computed using the jaw parameters $\theta$; each point cloud would only contain a range of jaw parameters and would therefore be smaller.
Moreover, it makes more sense to interpolate between shapes with similar jaw parameters as opposed to significantly different jaw parameters.
One should likely still unskin all of the shapes in the point cloud to have the same jaw parameter value for better efficacy; however, choosing a non-neutral reference shape for the unskinning (e.g.~in the middle of the relevant jaw parameter range) could be wise.

\section{Experiments}

\begin{figure}[b]
    \centering
    \includegraphics[width=\linewidth]{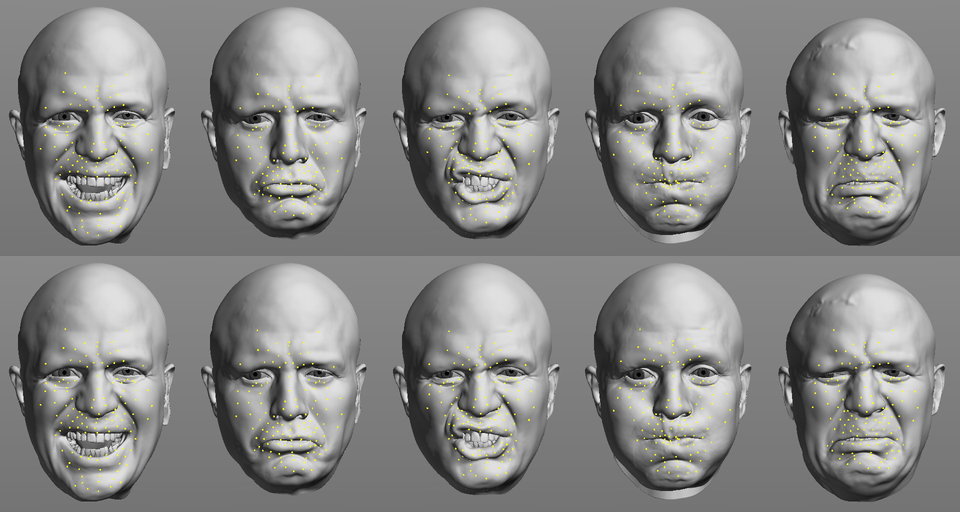}
    \caption{In order to verify our approach, we input 3D bundle positions from each shape in our library into our local geometric indexing algorithm; the results obtained are nearly identical to the original shapes. Top: Scan. Bottom: Local geometric indexing. A video showing the results on a larger dataset is available in the supplementary material. \label{fig:dataset_validation}}
\end{figure}

\begin{figure*}
    \centering
    \includegraphics[width=\linewidth]{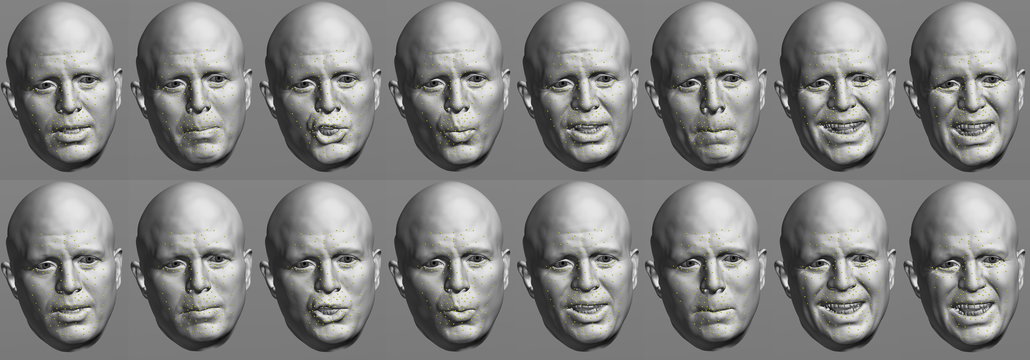}
    \caption{Top: A high-resolution facial performance processed using the Medusa performance capture system \cite{beeler2010high, beeler2014rigid, beeler2011high}. Bottom: Reconstruction obtained using local geometric indexing driven by the bundle positions on the captured geometry. None of the high-resolution facial shapes in the performance were included in the dataset used by our algorithm. A number of the differences, such as those in the mouth corners and eyebrows, are actually due to artifacts in the Medusa performance capture geometry that are cleaned up by our reconstruction indicating that our approach provides some degree of regularization. The remaining differences are outside of the region spanned by the bundles where we would expect less accuracy due to limited data. A video showing the results on the entire performance is available in the supplementary material. \label{fig:medusa_performance_validation}}
    \vspace{-0.2in}
\end{figure*}

\paragraph{Verification:} 
In order to verify our algorithm, we calculated a set of 3D bundles for each facial shape in our dataset by evaluating the surface position of each bundle on the facial shape.
Then, we inputted each set of bundle positions into our local geometric indexing algorithm, and verified that the resulting reconstruction is nearly identical to the original facial shape.
See Figure~\ref{fig:dataset_validation}.

\vspace{-0.1in}
\paragraph{High-Resolution Capture Comparison:} 
Next, we evaluate our algorithm on a high-resolution performance outputted from the Medusa performance capture system \cite{beeler2010high, beeler2014rigid, beeler2011high}.
The jaw is tracked using the lower teeth during the portions of the performance where they are visible and interpolated to the rest of the performance using the chin bundles as a guide.
Like the previous experiment, we calculate a set of 3D bundles for each frame of the performance and use this animated set of 3D bundles as input into our local geometric indexing algorithm.
The resulting high-resolution reconstruction of the performance using our dataset is very similar to the original performance.
See Figure~\ref{fig:medusa_performance_validation}.
The differences in the mouth corners and lips are due to artifacts in the Medusa performance capture.
By indexing the most relevant cleaned up shapes in our dataset, we obtain a cleaner reconstruction while also adding detail sculpted by a modeler such as lip wrinkles.
Other differences, such as those on the forehead and side of the face, occur because there are no bundles in those locations and thus our algorithm extrapolates from the nearest bundle.

\begin{figure*}
    \centering
    \fivefigure{.195\linewidth}{.01in}{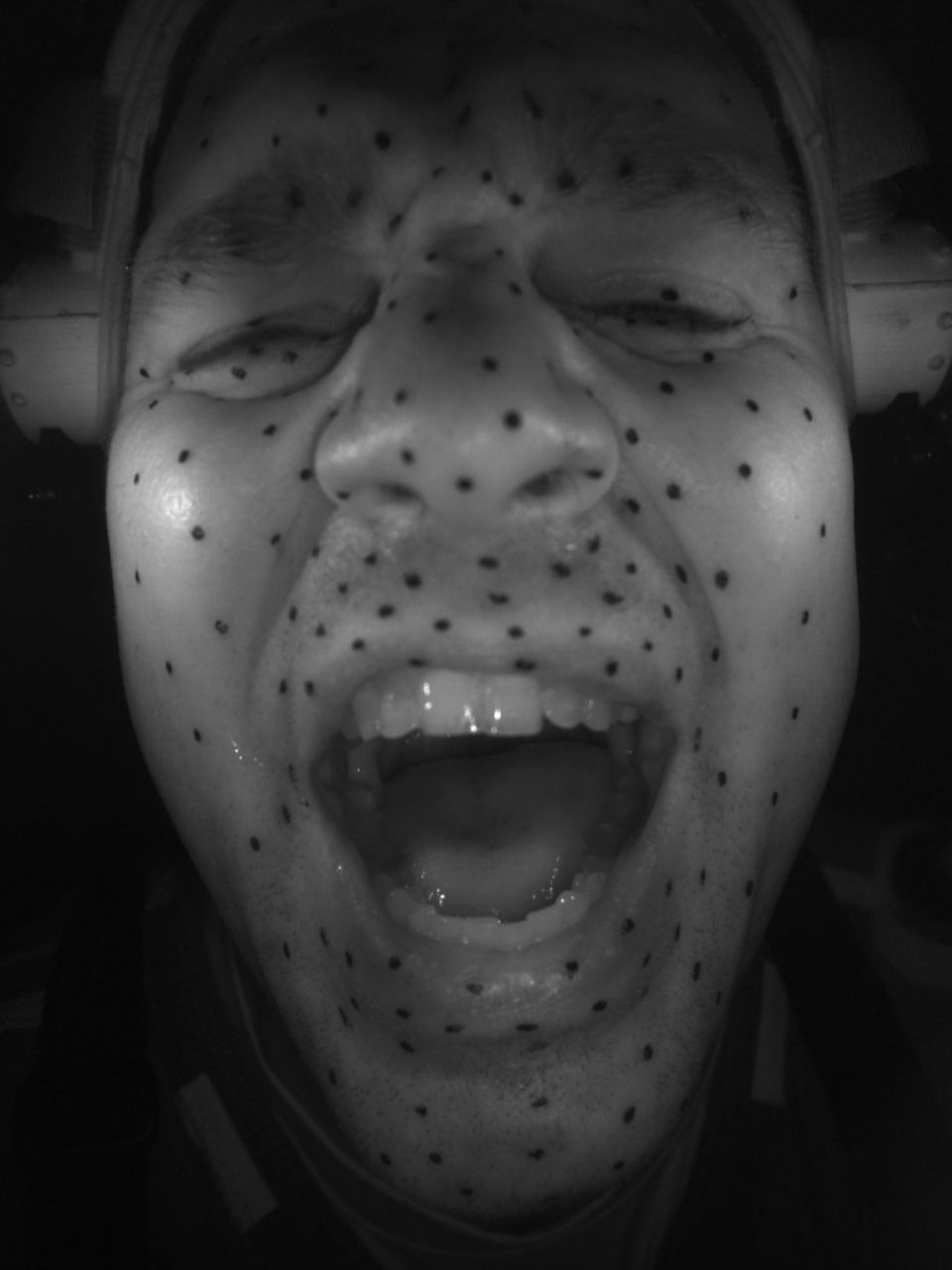}{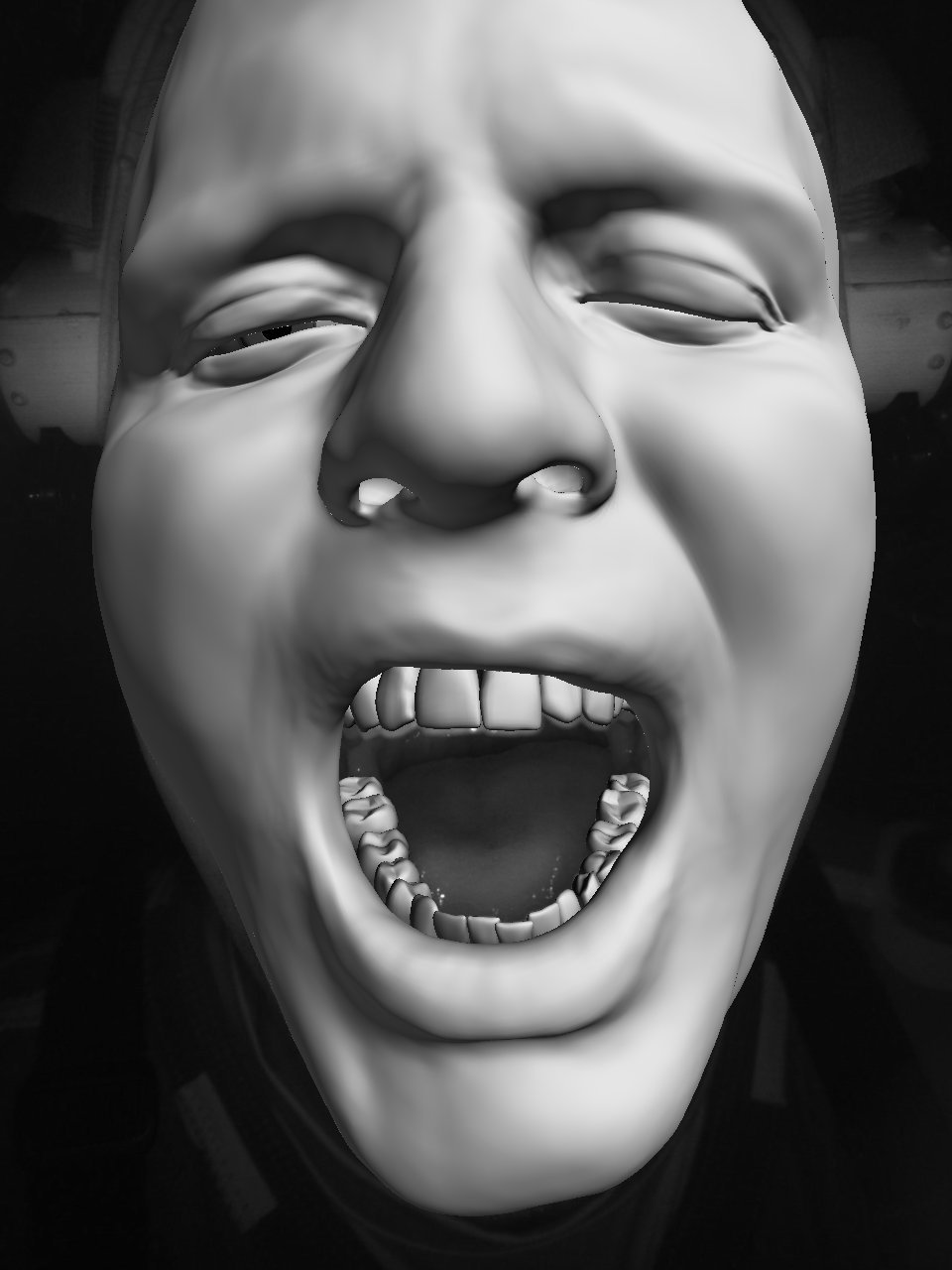}{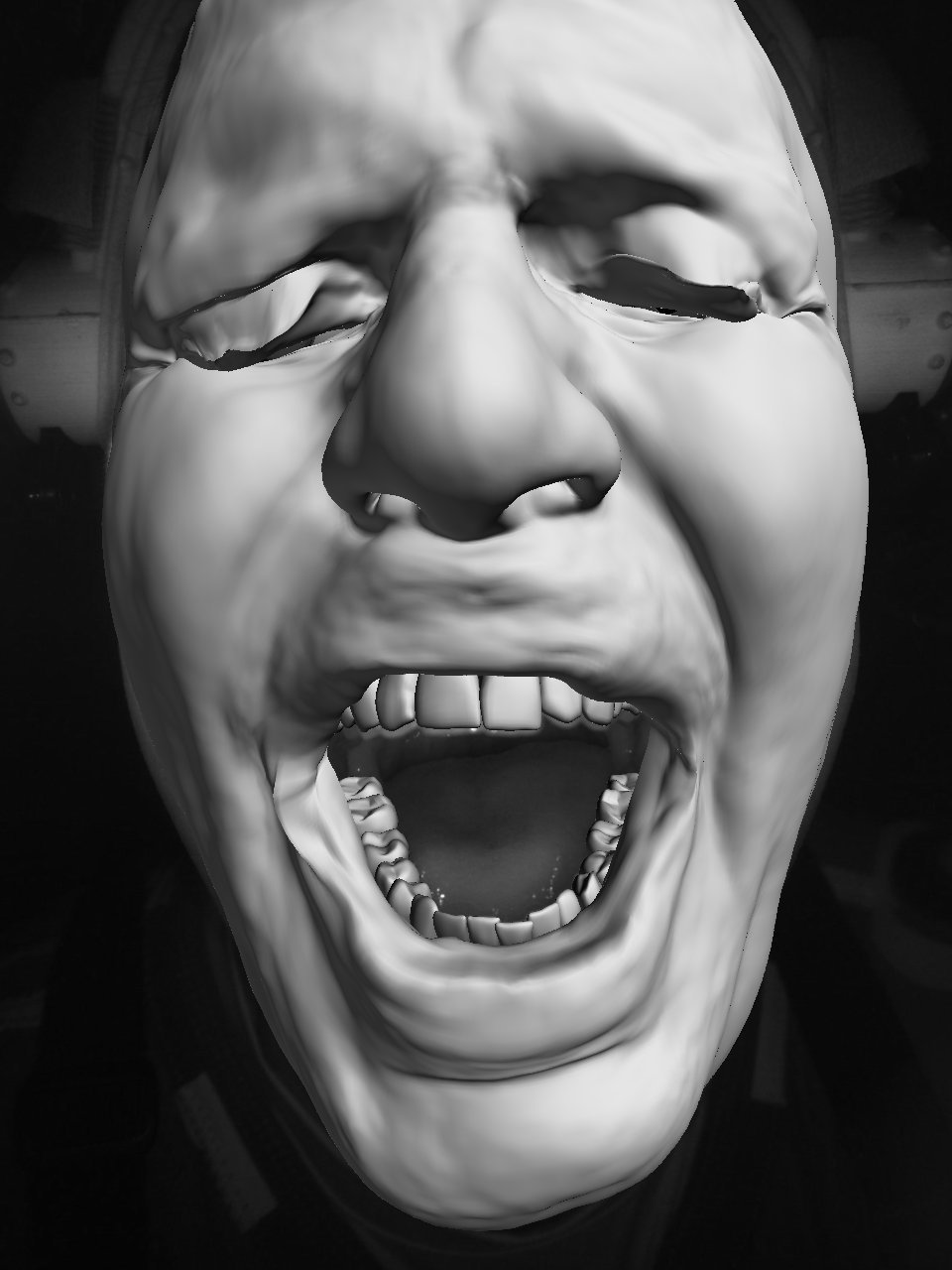}{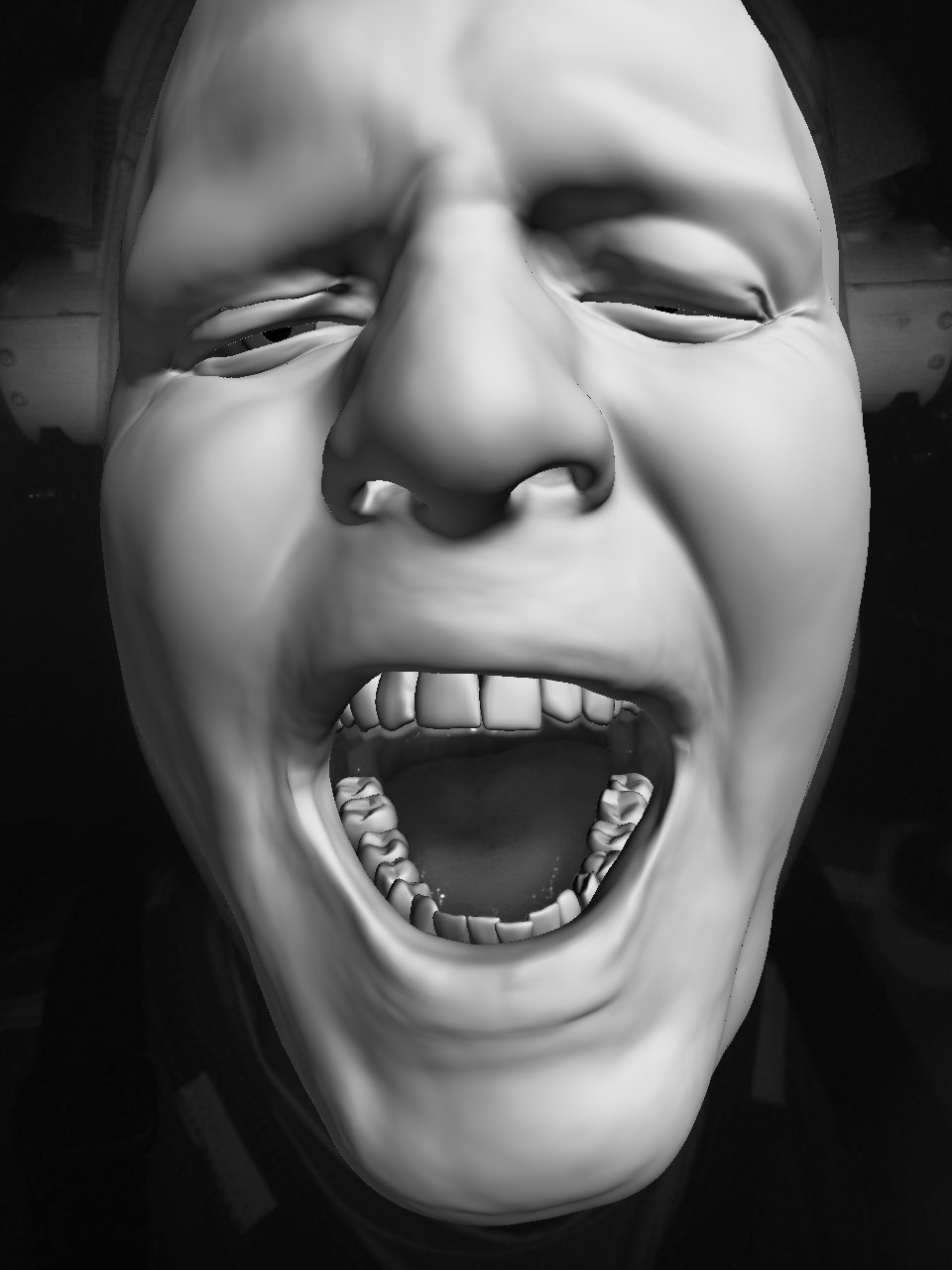}{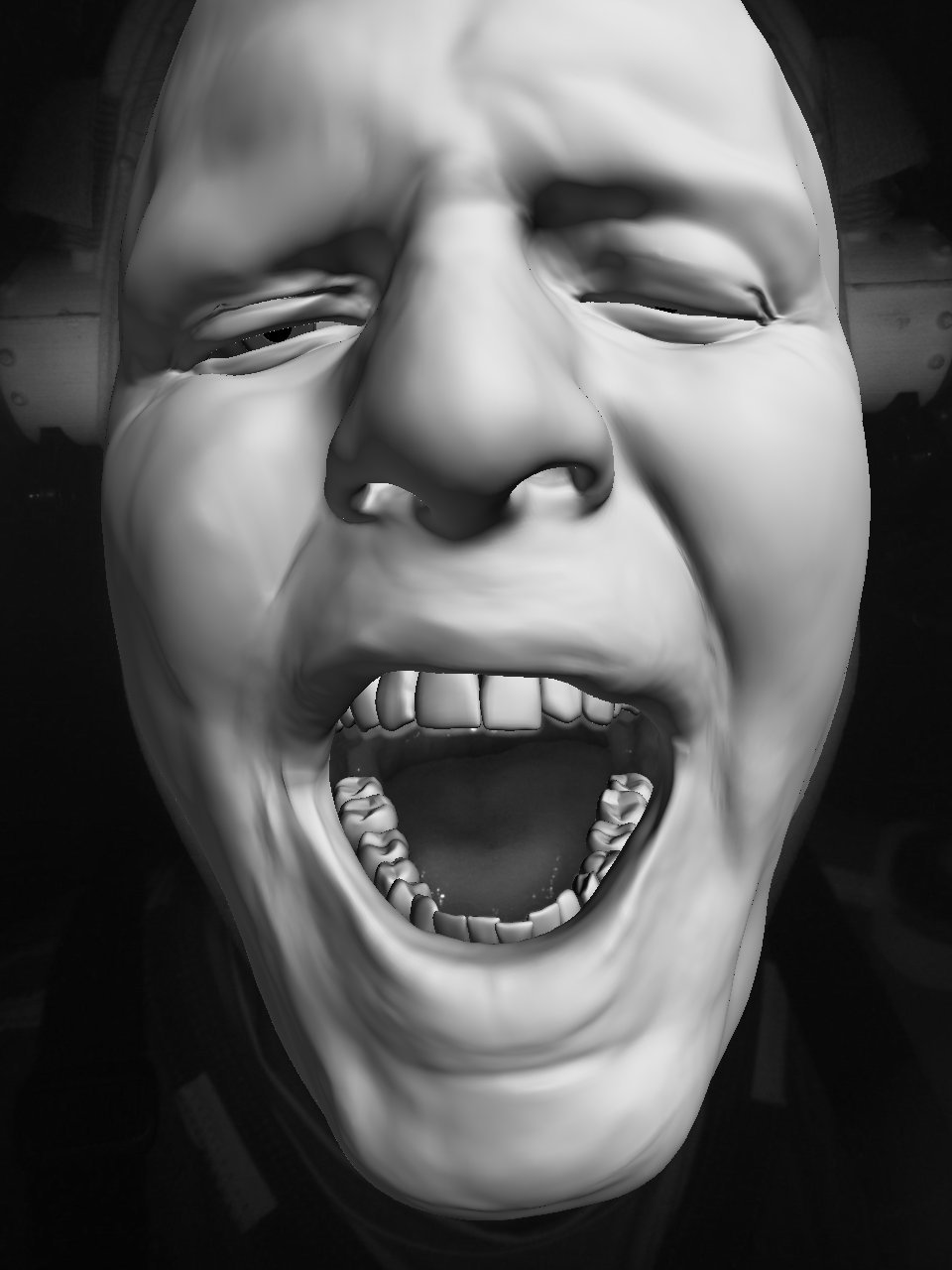}
    \caption{Far Left: Helmet mounted camera footage. Middle Left: The reconstruction obtained by interpolating the bundle displacements across the mesh using \cite{bhat2013high} conveys a yawn as opposed to the anger/tension because it does not utilize any additional high-resolution detail beyond that of the neutral mesh. Middle: The typical overfitting symptomatic of blendshape rigs; with enough regularization, one would expect the detail to fade similar to the result using \cite{bhat2013high}. Middle Right: Using Gaussian RBF interpolation instead of natural neighbor interpolation in our approach results in additional high-resolution detail but does not interpolate the bundle positions. Far Right: Our approach passes through the bundles, conveys the expression, and captures high-resolution details that are not present in the neutral mesh. \label{fig:hmc_comparison}}
    \vspace{-0.1in}
\end{figure*}

\vspace{-0.1in}
\paragraph{Comparison to Other Approaches:} 
In Figure~\ref{fig:hmc_comparison}, we compare our approach to other popular approaches on a performance captured using two vertically stacked helmet mounted fisheye cameras.
Footage from the top camera placed at nose level is shown in Figure~\ref{fig:hmc_comparison} (Far Left).
The images from both cameras are undistorted and the cameras are calibrated using the markers on the helmet.
The calibrated cameras are used to triangulate bundle positions which are then rigidly aligned to the neutral mesh using a combination of the bundles on the nose bridge, forehead, and the cheeks with varying weights based on the amount of non-rigid motion in those regions.
The jaw is tracked in the same manner as the previous experiment.
As shown in Figure~\ref{fig:hmc_comparison} (Middle Left), interpolating the bundle displacements across the mesh using \cite{bhat2013high} reconstructs a yawn instead of the angry face in the corresponding helmet mounted camera footage because it does not contain any additional high-resolution detail beyond that of the neutral mesh.
Since the neutral mesh represents one's face while expressionless, similar to that when asleep, using the displacements of the neutral mesh and its features often leads to expressions that appear tired.
In order to obtain Figure~\ref{fig:hmc_comparison} (Middle), we first constructed a blendshape rig using the facial shapes in our dataset.
Then, we solved for the blendshape weights that minimize the Euclidean distances from the bundles to their relevant surface points subject to a soft constraint that penalizes the weights to lie between 0 and 1.
The result incorporates more high-resolution details than Figure~\ref{fig:hmc_comparison} (Middle) but suffers from overfitting resulting in severe artifacts around the mouth and eyes.
Even though the resulting weights lie between 0 and 1, they are neither convex nor sparse which leads to unnatural combinations.
Of course, increased regularization would smooth the artifacts shown in the figure creating a result that looks more like Figure~\ref{fig:hmc_comparison} (Middle Left).
In comparison, the reconstruction obtained using our local geometric indexing algorithm shown in Figure~\ref{fig:hmc_comparison} (Far Right) captures many of the high-resolution details that are not present in the neutral mesh including the deepened nasolabial folds, jowl wrinkles, and lip stretching without the overfitting artifacts of Figure~\ref{fig:hmc_comparison} (Middle).

\begin{figure*}[b]
    \centering
    \fourfigure{.23\linewidth}{.01in}{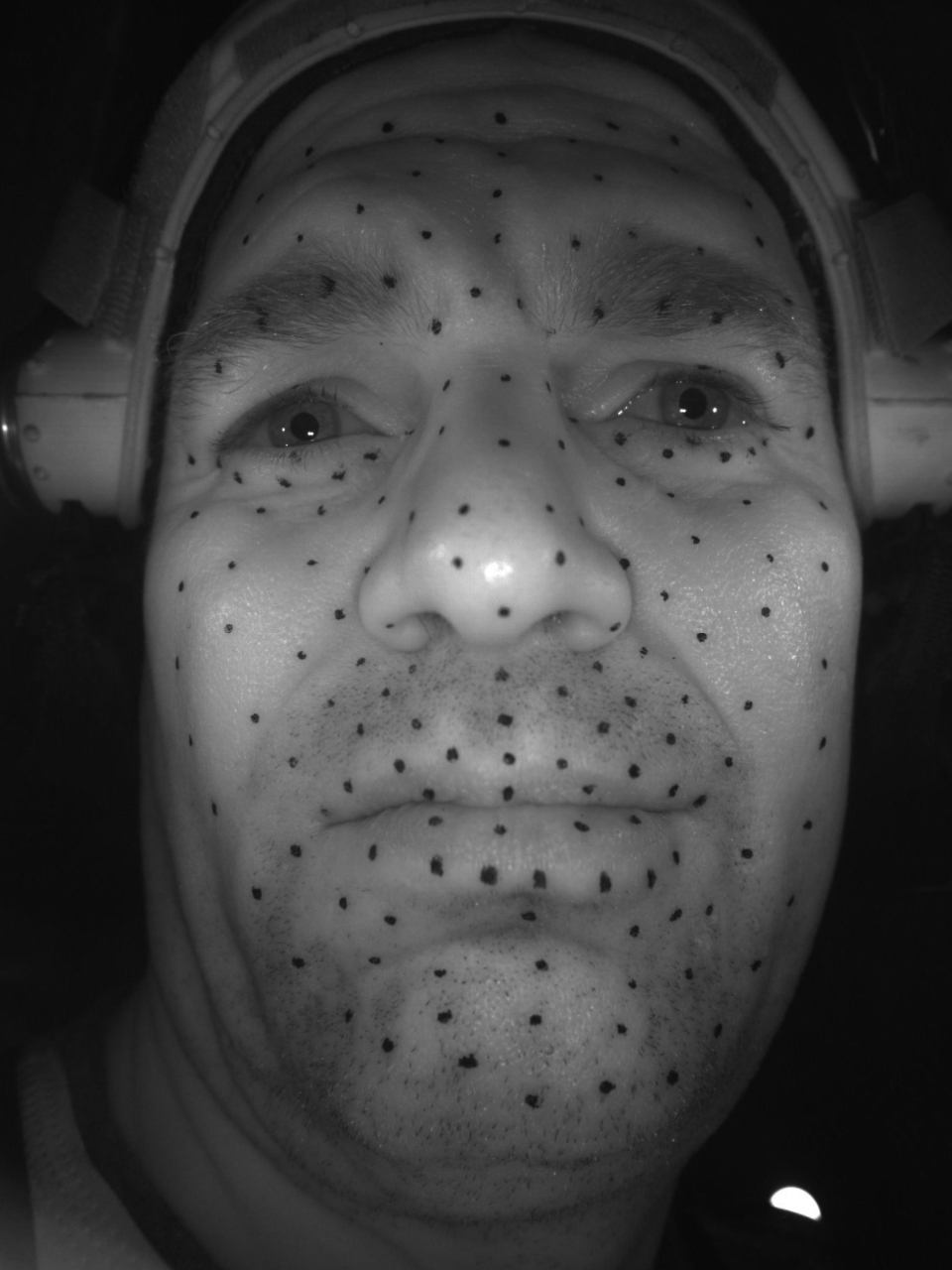}{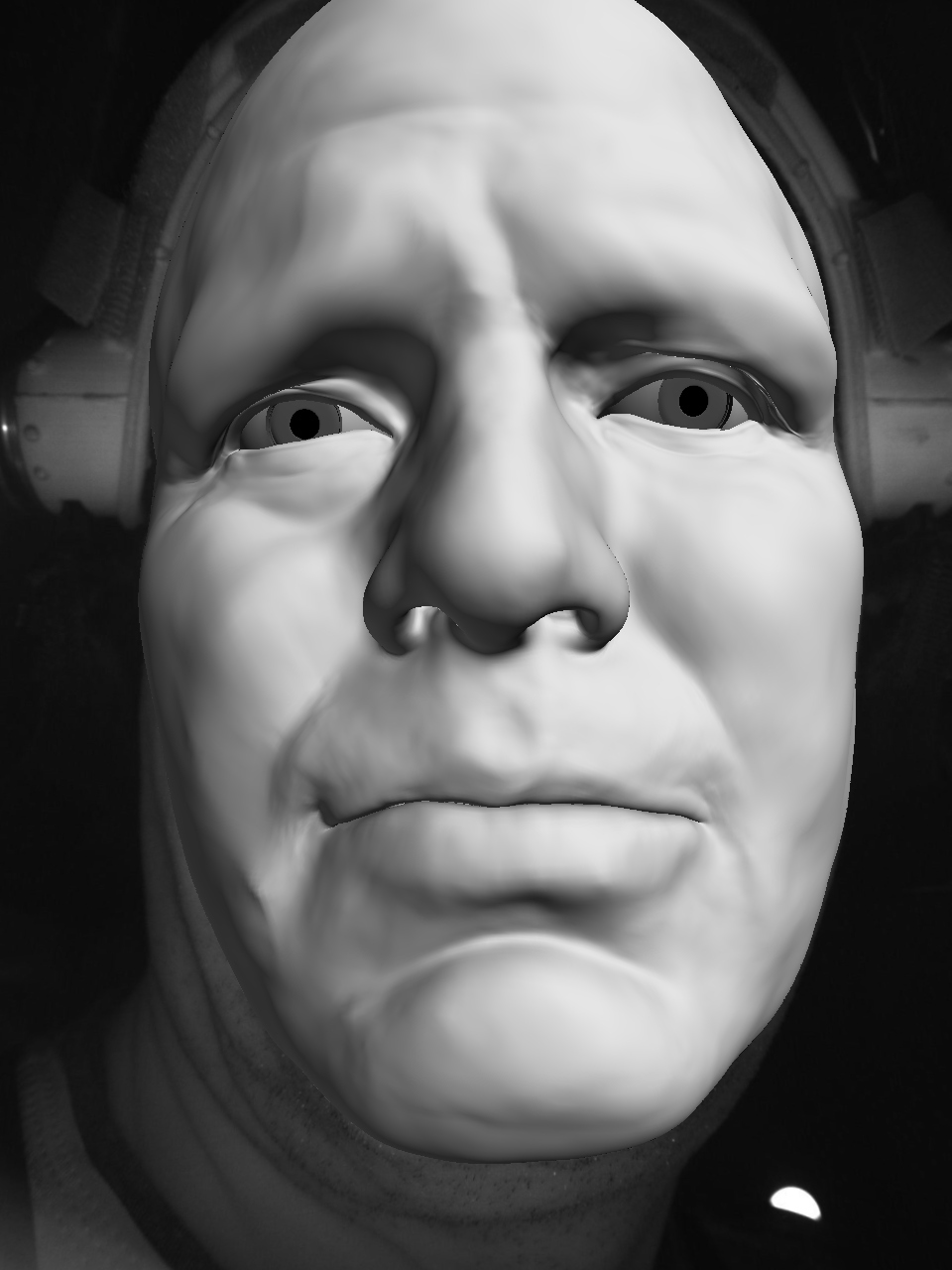}{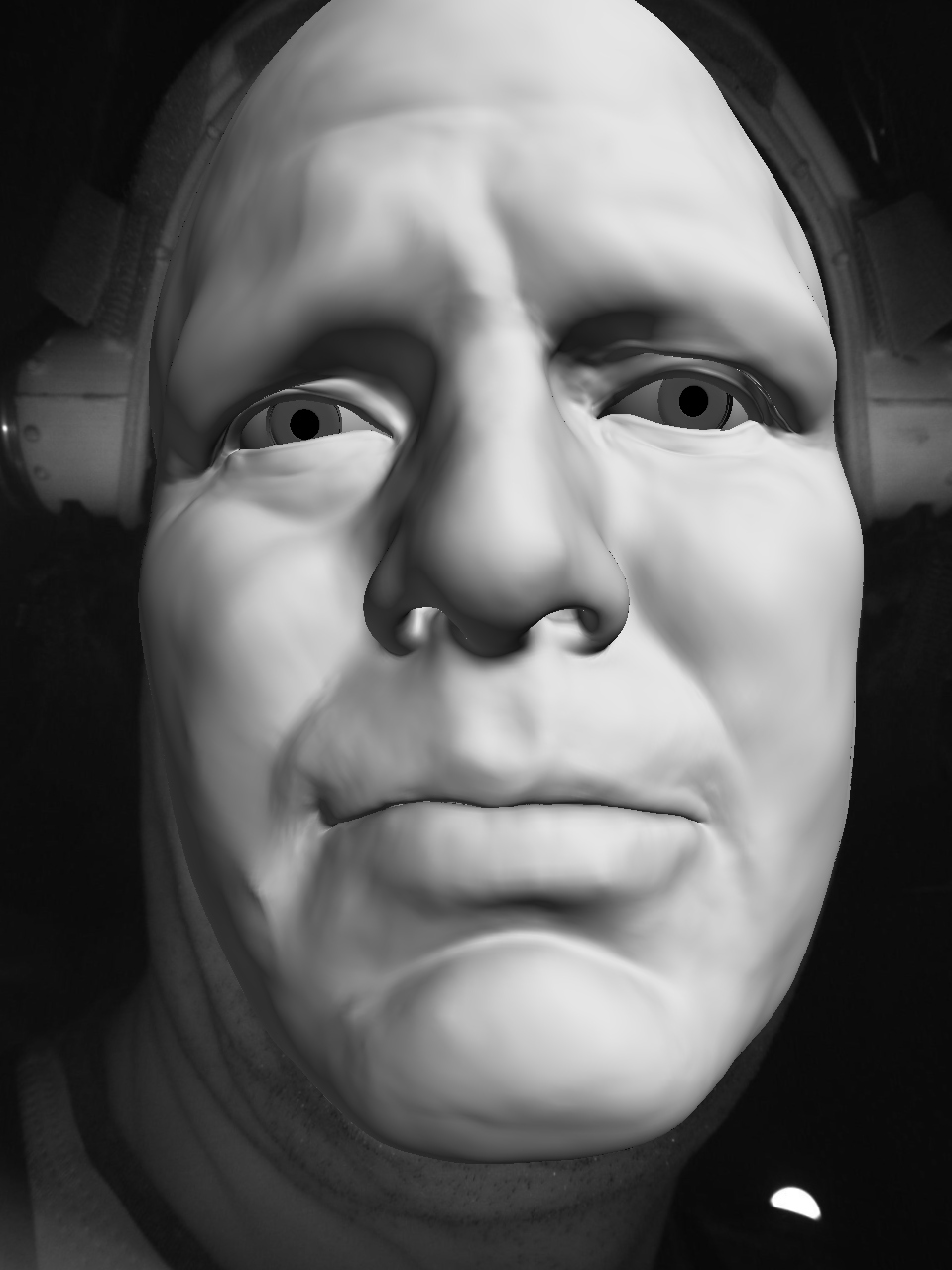}{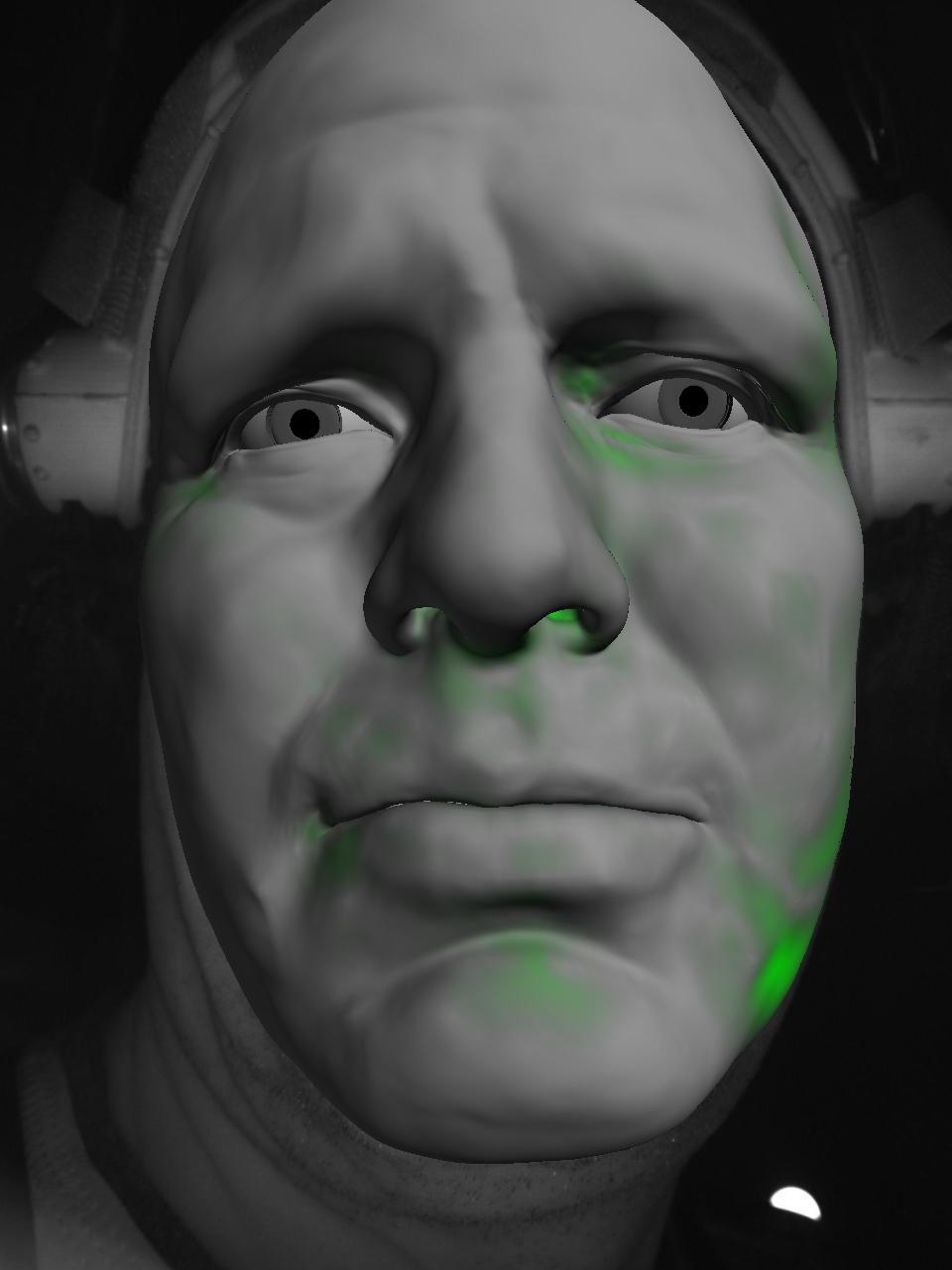}
    \caption{Far Left: Helmet mounted camera footage. Middle Left: The reconstruction obtained using our approach captures a subtle expression in the helmet mounted camera footage. This performance also shows the effectiveness of our temporal smoothness constraints. See video in supplementary material. Middle Right: Adding simulated in-betweens allows us to improve the smoothness of the reconstruction in the philtrum and the right jowl while also improving the lift in the upper right cheek. Far Right: Heatmap highlighting the differences between (Middle Left) and (Middle Right). \label{fig:hmc_temporal_smoothness}}
\end{figure*}

\vspace{-0.1in}
\paragraph{RBF Interpolation:} 
Alternatively, instead of using natural neighbor interpolation, one could use radial basis functions to smooth with our local geometric indexing algorithm.
As long as the radial basis function is applied on the facial shape weights as opposed to the vertex positions themselves, this still yields high-resolution features from the dataset in the reconstruction; however, the reconstructed surface will typically not pass through the bundles.
This can be corrected by smoothly interpolating the remaining displacements needed to properly interpolate the bundles across the mesh (with e.g.~\cite{bhat2013high}).
As shown in Figure~\ref{fig:hmc_comparison} (Middle Right), the reconstruction obtained using a combination of radial basis function interpolation and a smoothly interpolated deformation has a higher degree of detail than smoothly interpolating the deformation from neutral mesh.
In a similar manner, additional rotoscoped constraints such as lip occlusion contours \cite{bhat2013high, dinev2018user}, markers visible in only a single camera, etc. can be incorporated as a postprocessing step on top of our approach; in fact, we utilized \cite{bhat2013high} to incorporate lip occlusion contours in Figure~\ref{fig:hmc_comparison}.

\begin{figure*}
    \centering
    \sixfigure{.166666\linewidth}{.01in}{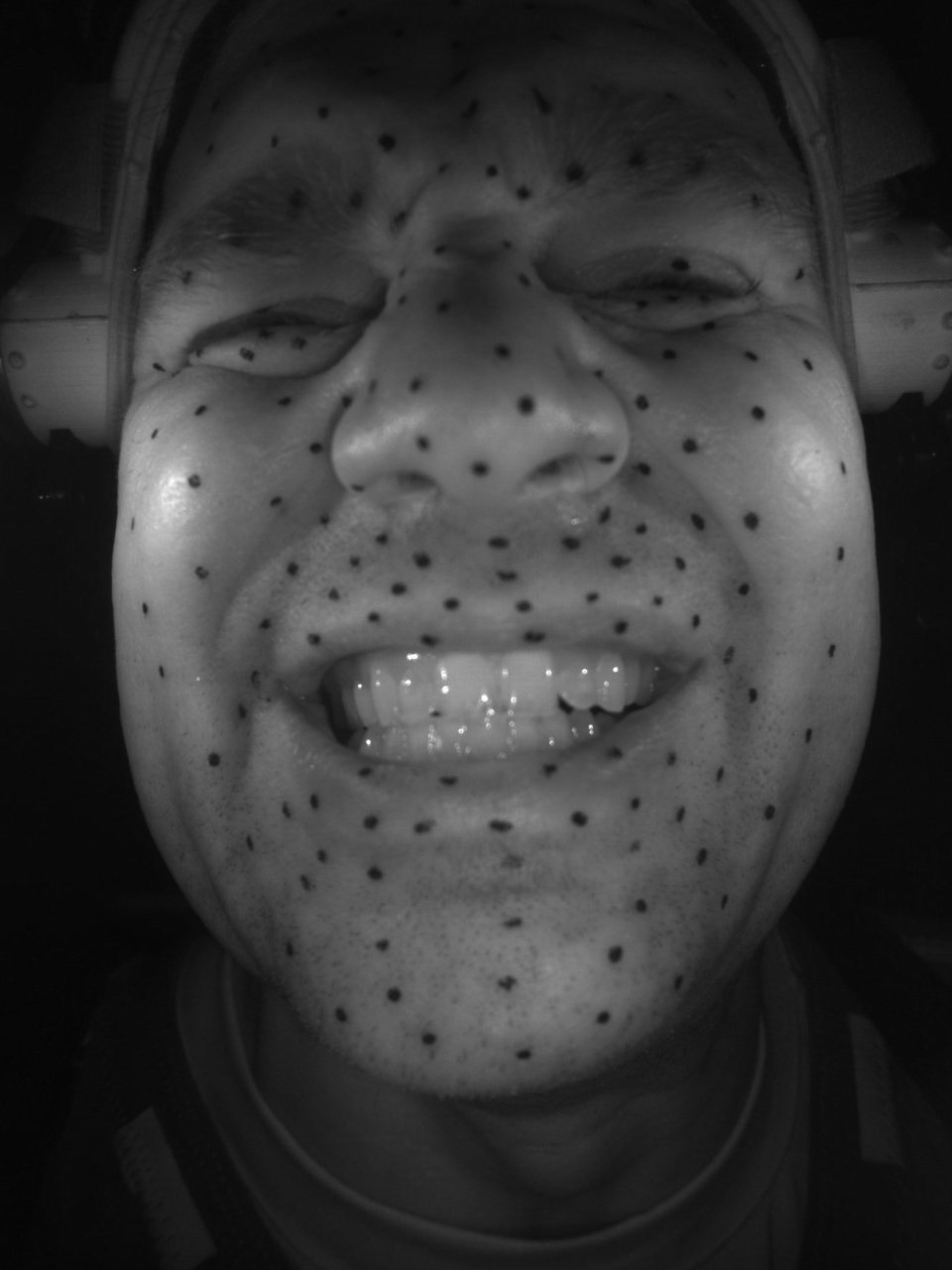}{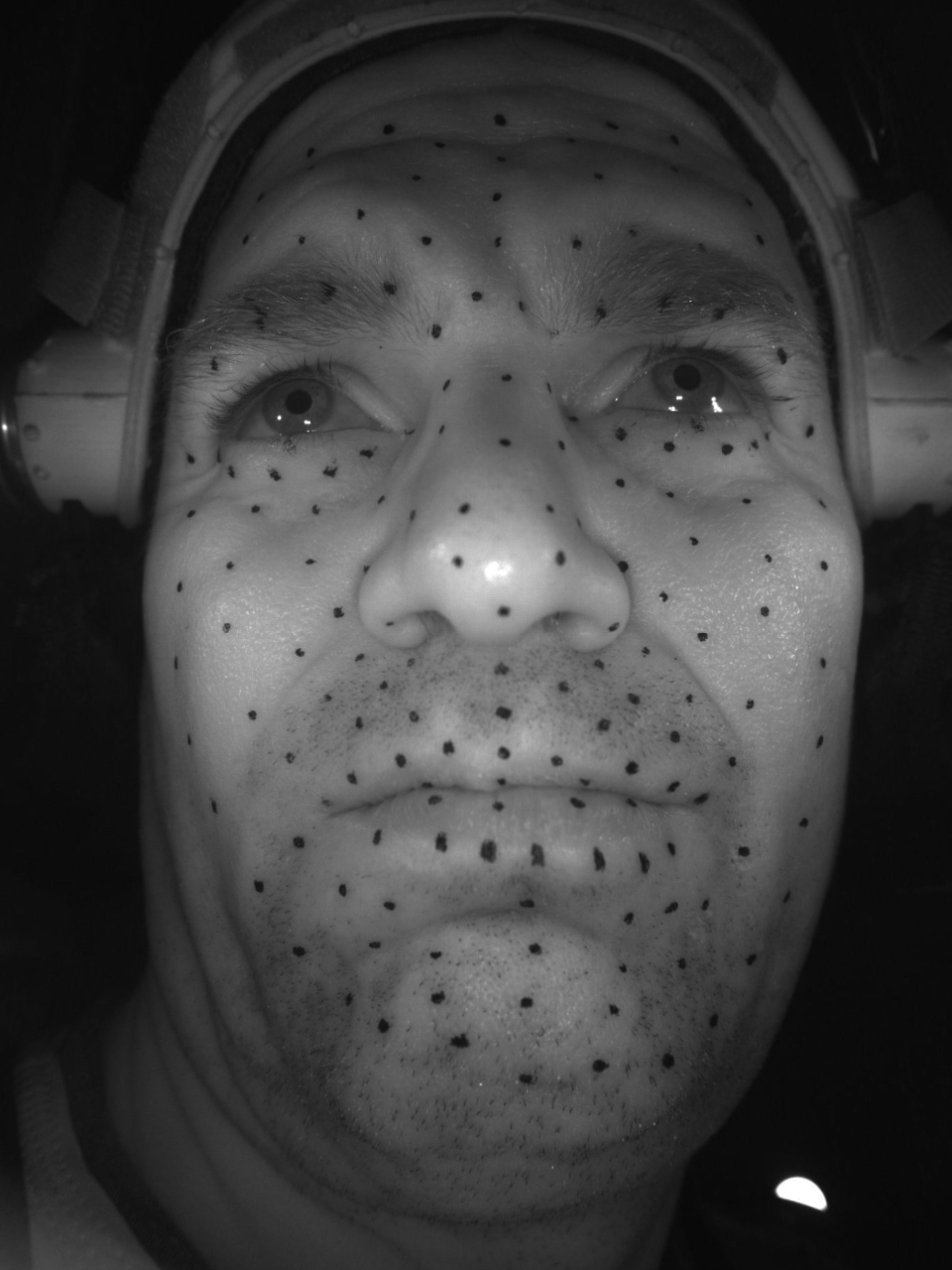}{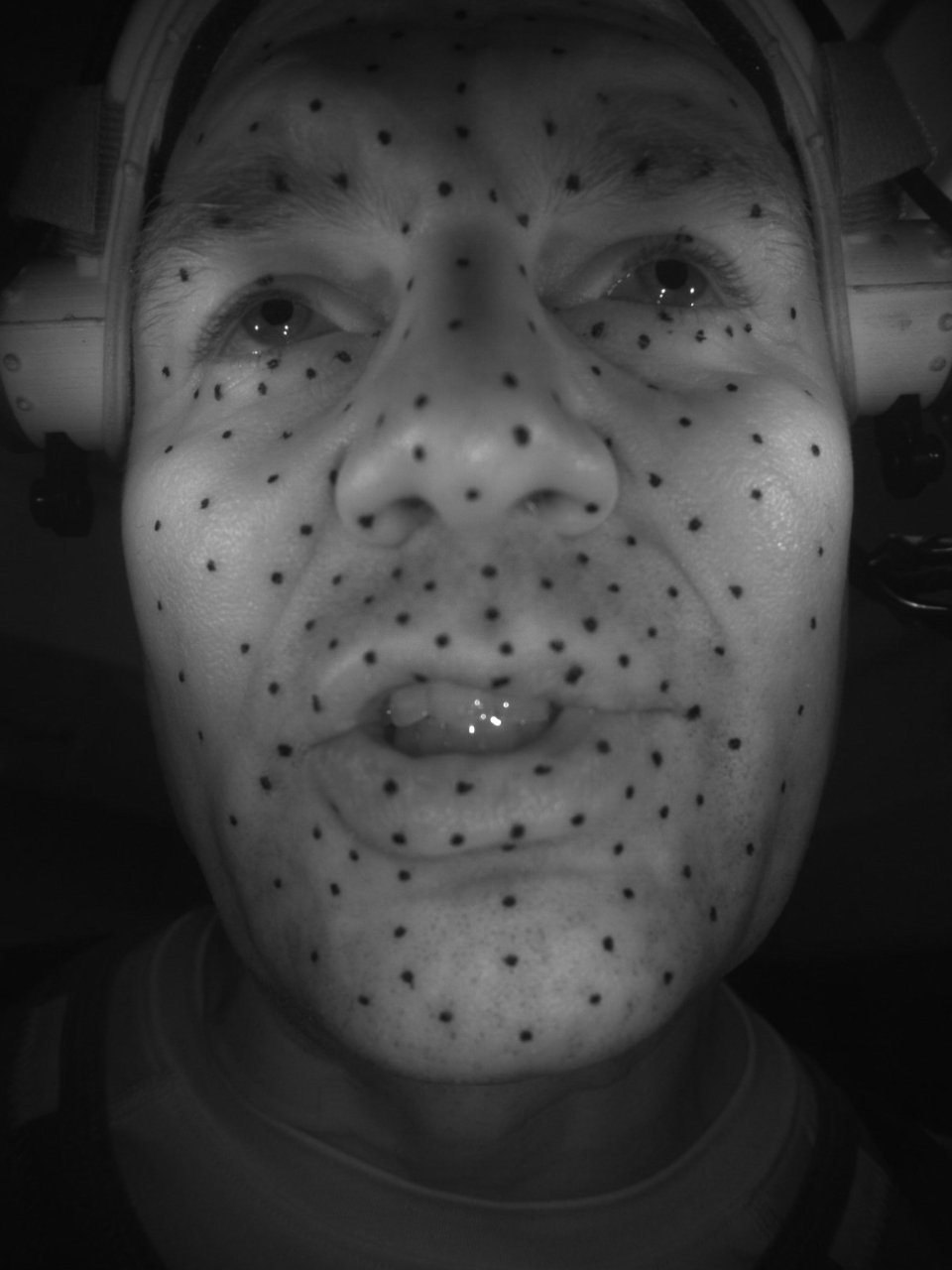}{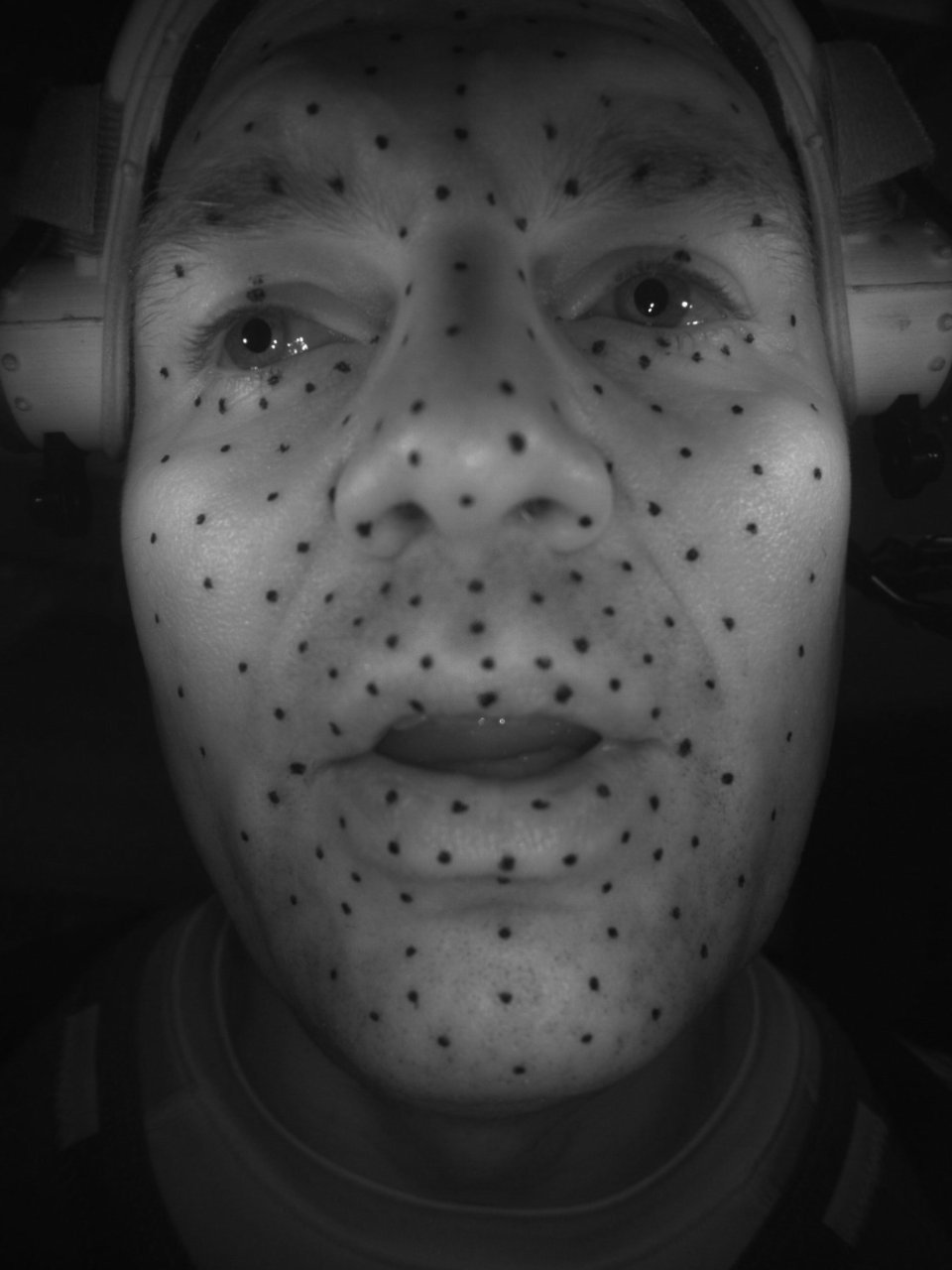}{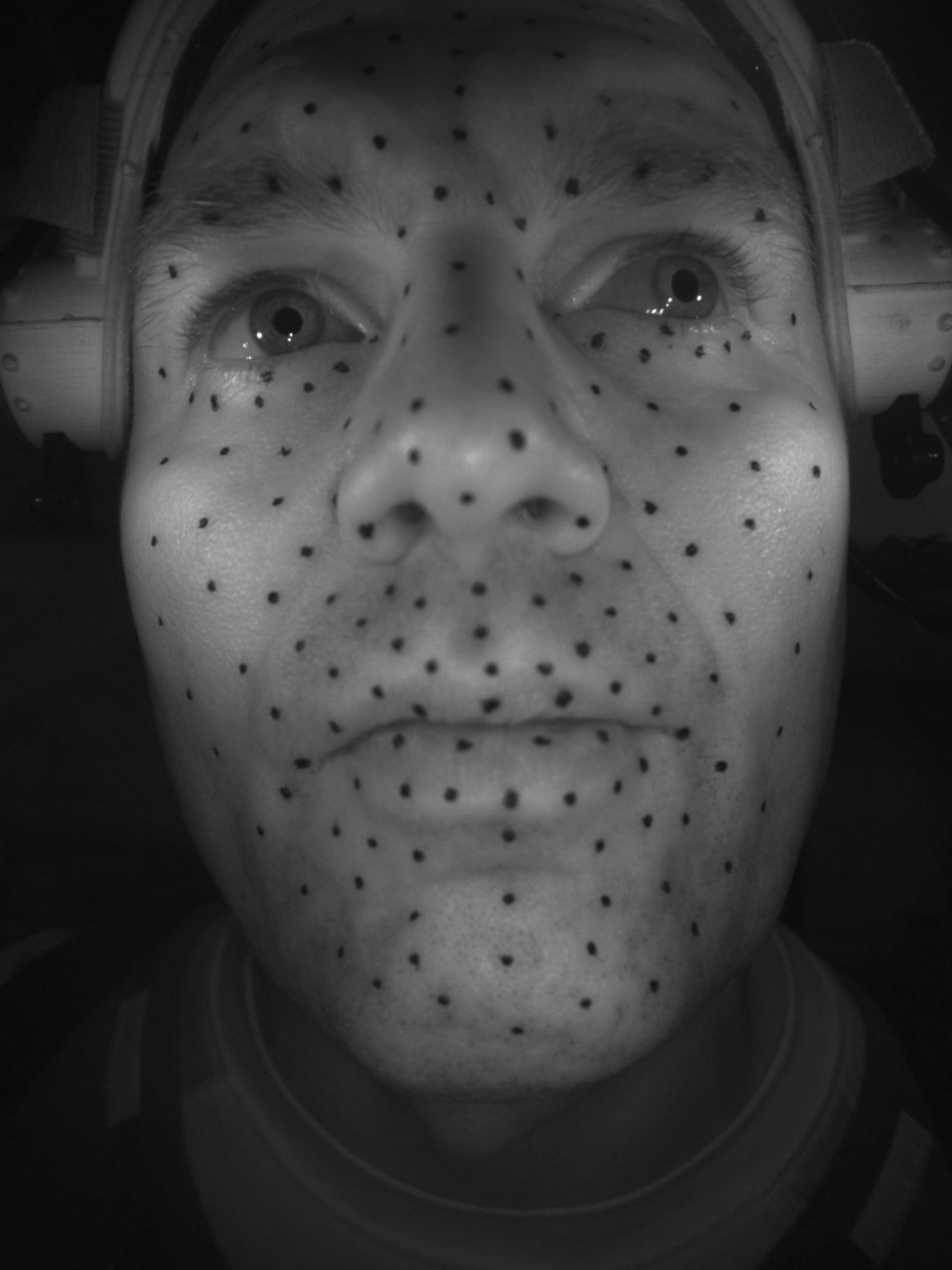}{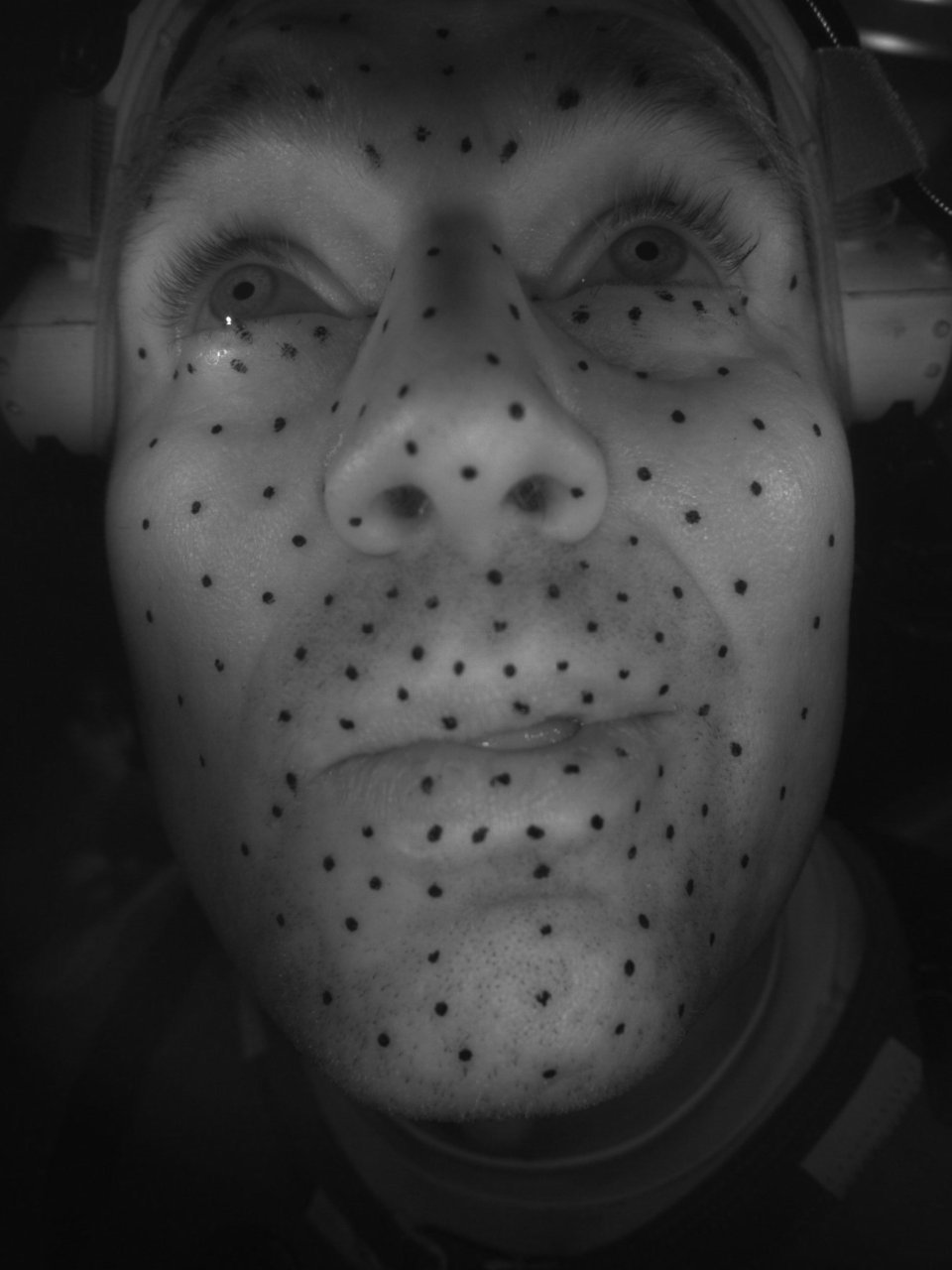}
    \sixfigure{.166666\linewidth}{.01in}{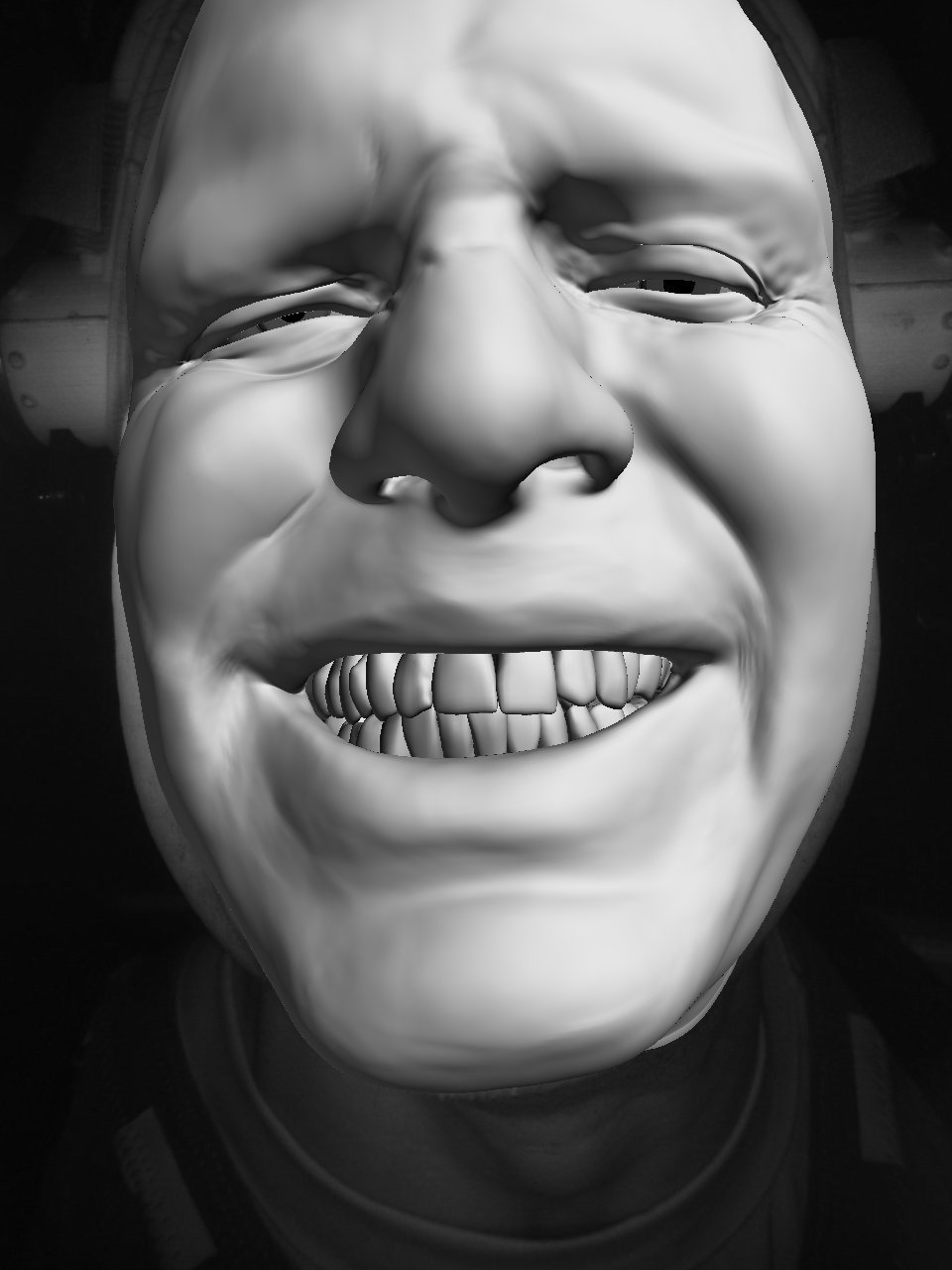}{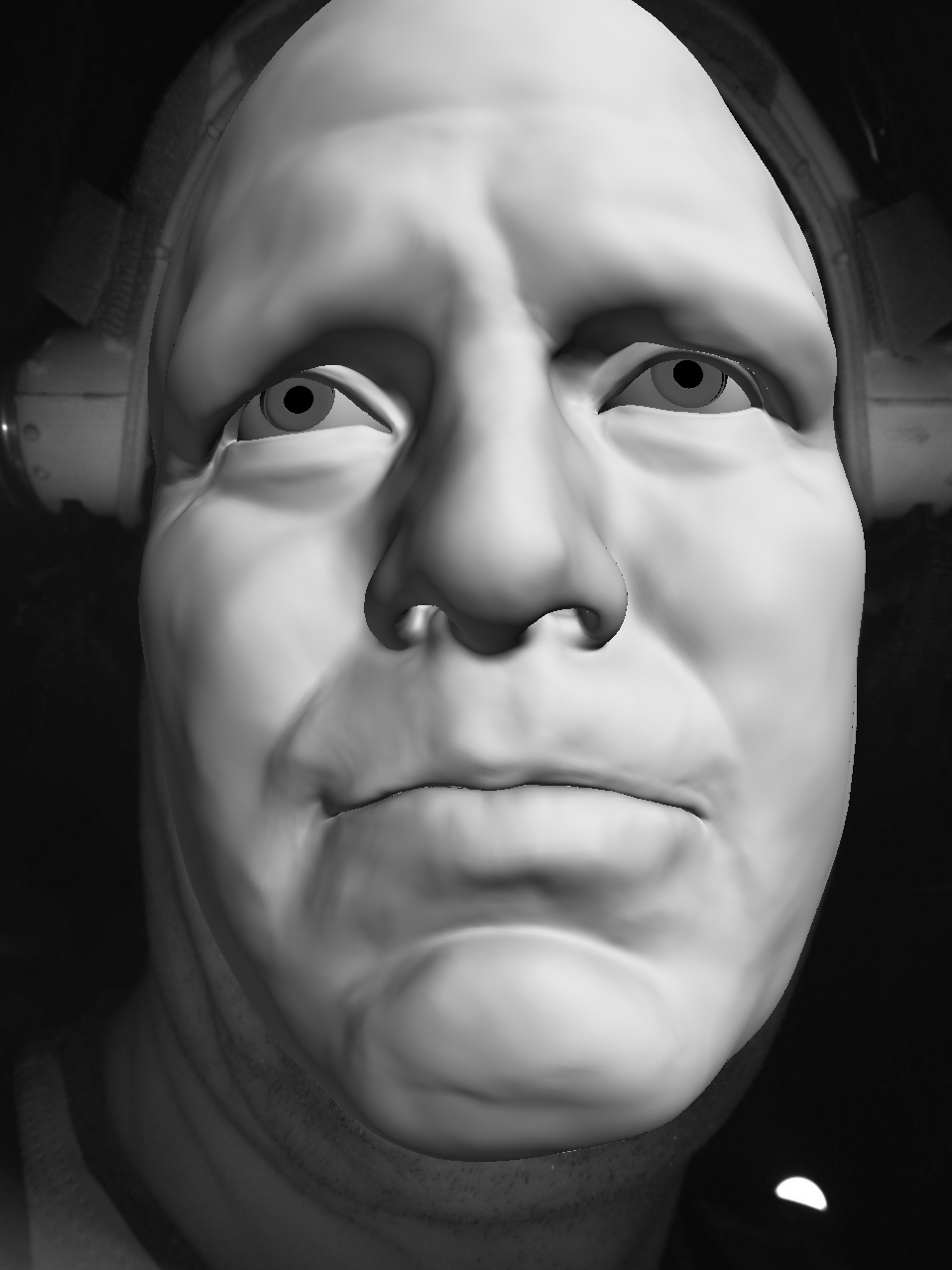}{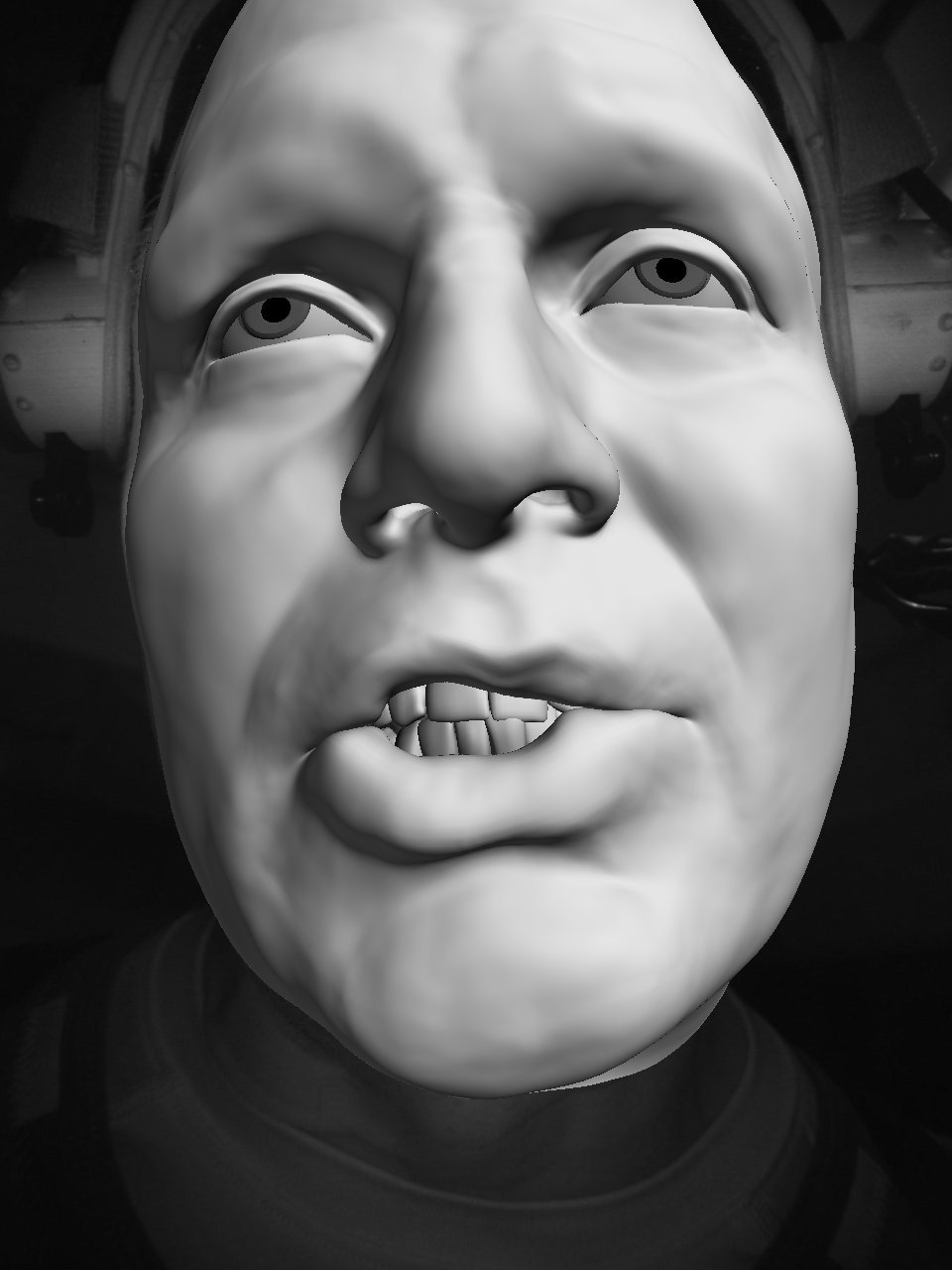}{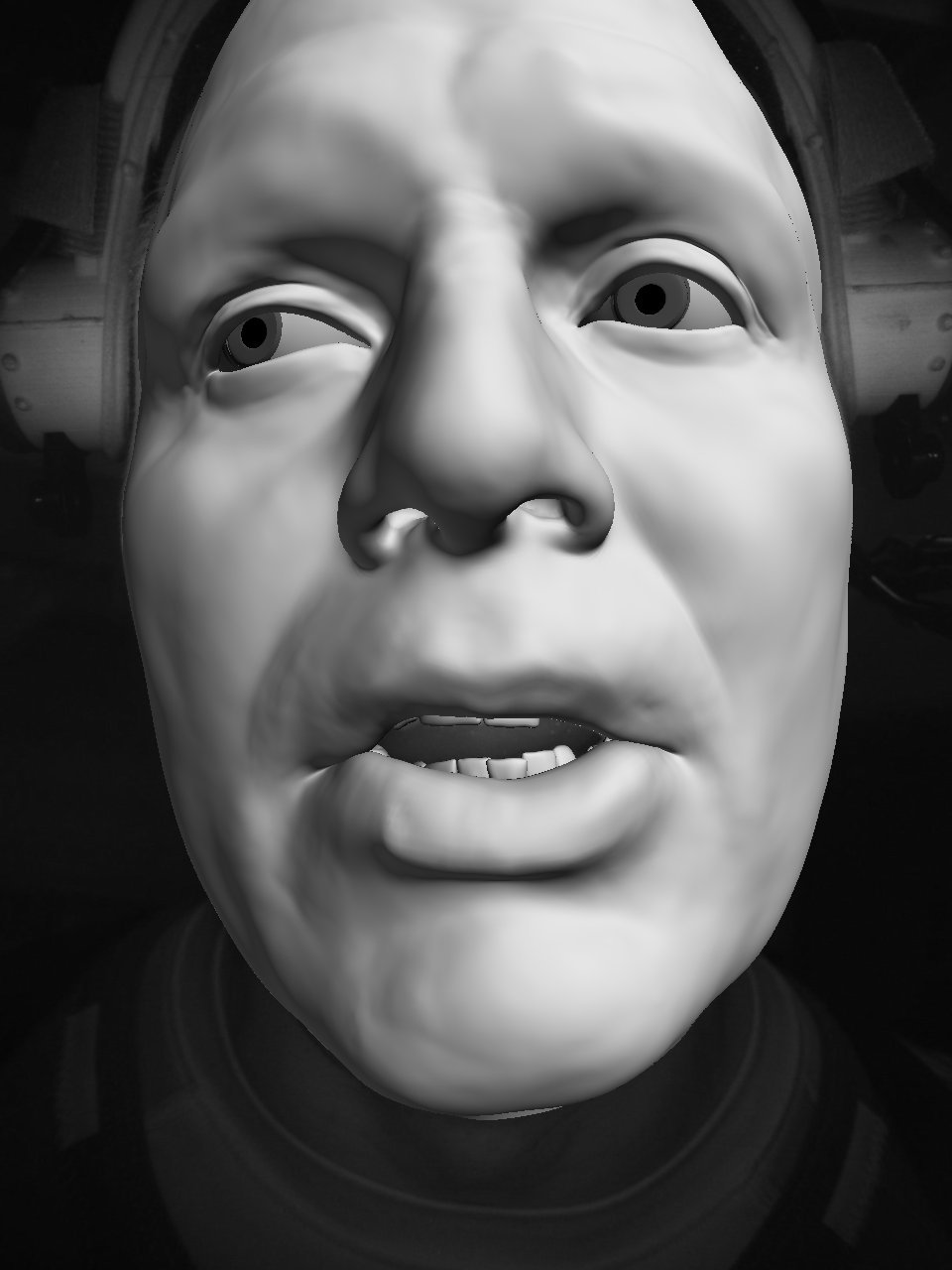}{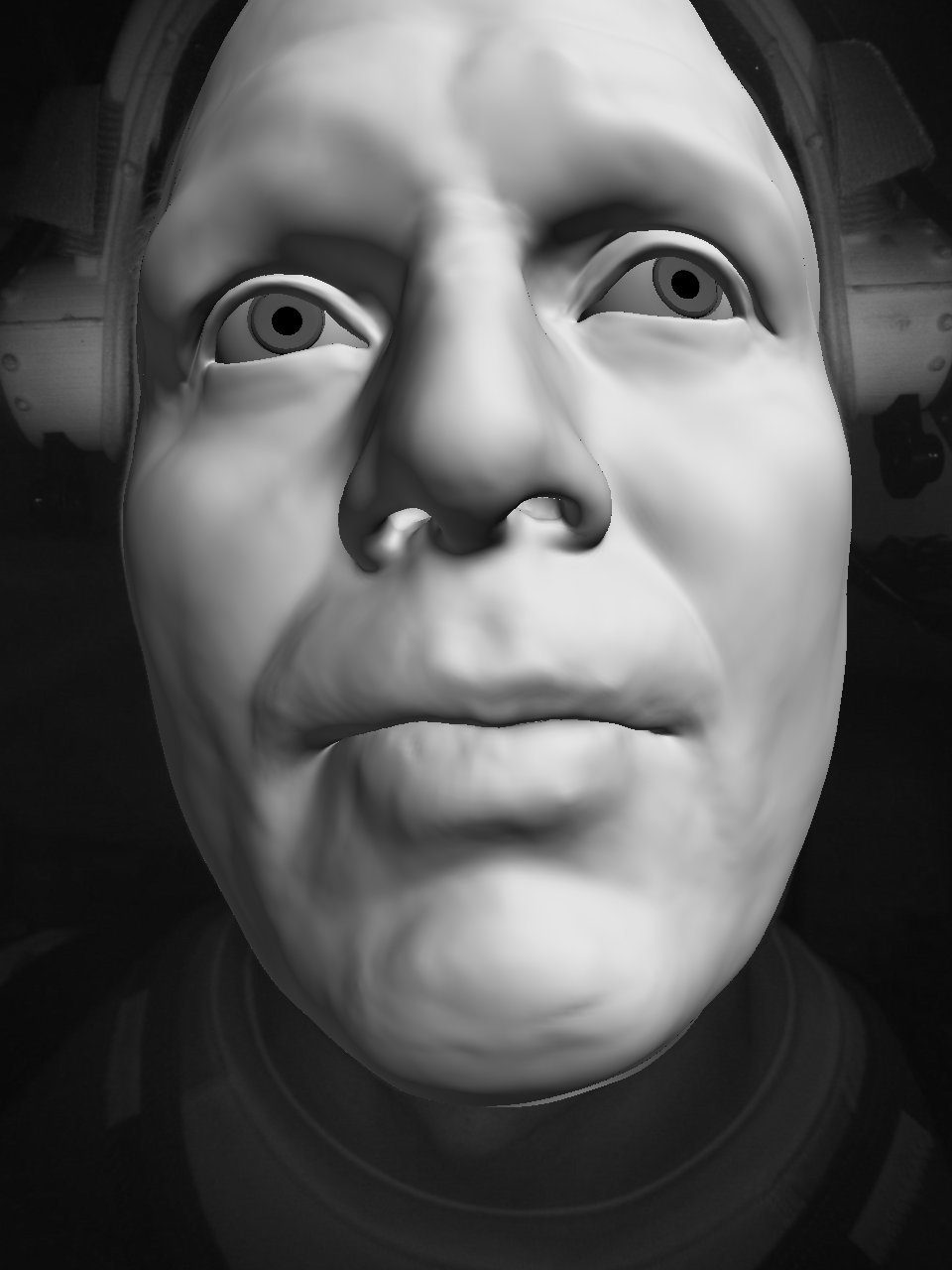}{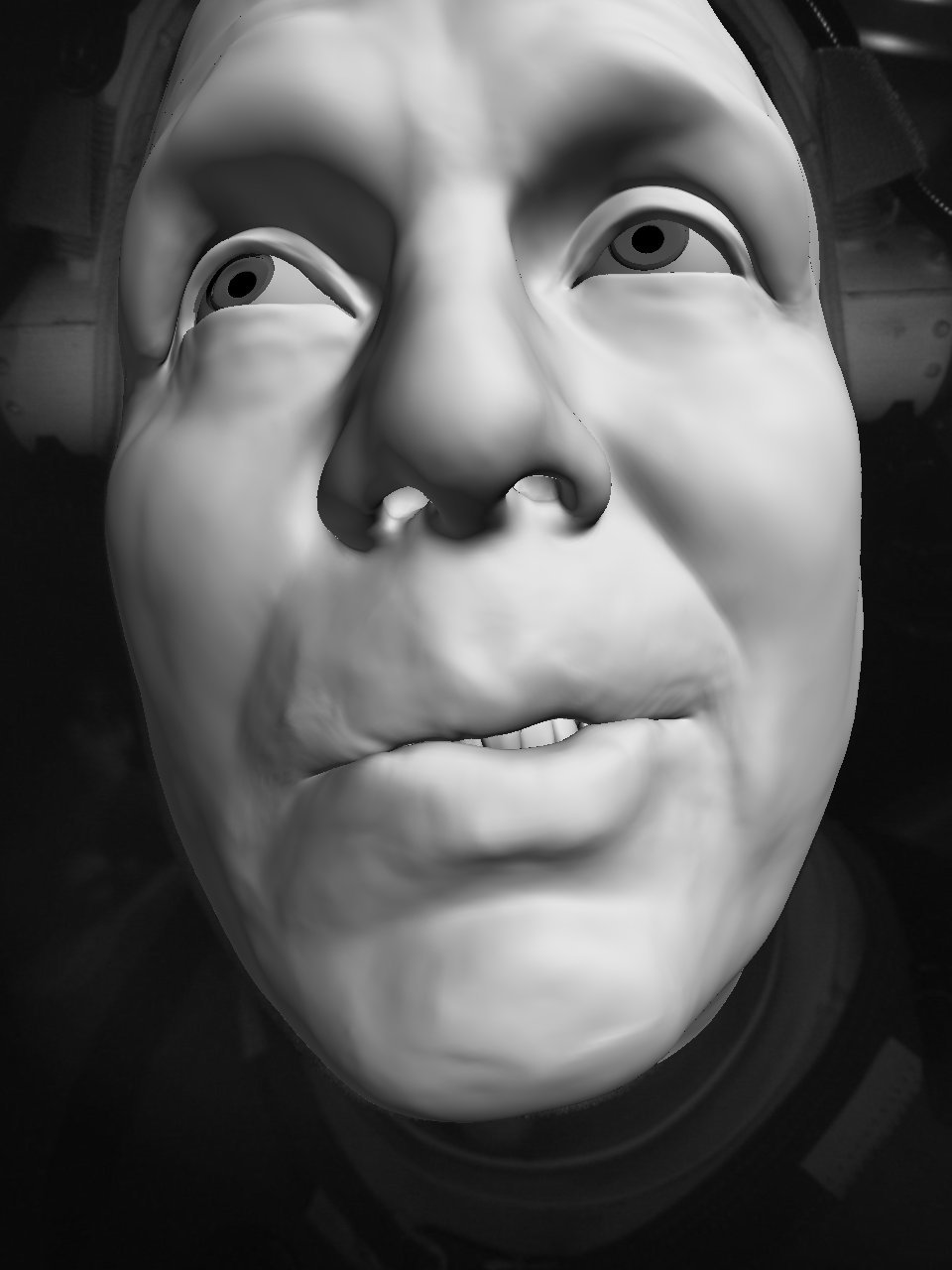}
    \caption{Top: Helmet mounted camera footage. Bottom: Reconstructions obtained using our method. \label{fig:hmc_examples}}
    \vspace{-0.2in}
\end{figure*}

\vspace{-0.1in}
\paragraph{Temporal Smoothing:} 
Figure~\ref{fig:hmc_temporal_smoothness} demonstrates the ability of our method to capture subtle expressions while also maintaining temporal coherency in the presence of bundle positions with random and systematic errors (e.g.~errors in depth due to the limited parallax between the two cameras).
If necessary, one can obtain a smoother performance by either temporally smoothing the input bundle positions or smoothing the barycentric weights on each bundle.
In this performance, we apply temporal smoothing by taking a central moving average of the barycentric weights associated with each bundle relative to the jaw skinned neutral mesh in order to avoid smoothing the jaw animation.
Because transitions between different sets of shapes typically occur when the same bundle position is achievable using multiple tetrahedra, we found this straightforward temporal smoothing scheme to have negligible impact on the ability for the reconstruction to interpolate the bundles.

\begin{figure}[b]
    \centering
    \twofigure{.495\linewidth}{.03in}{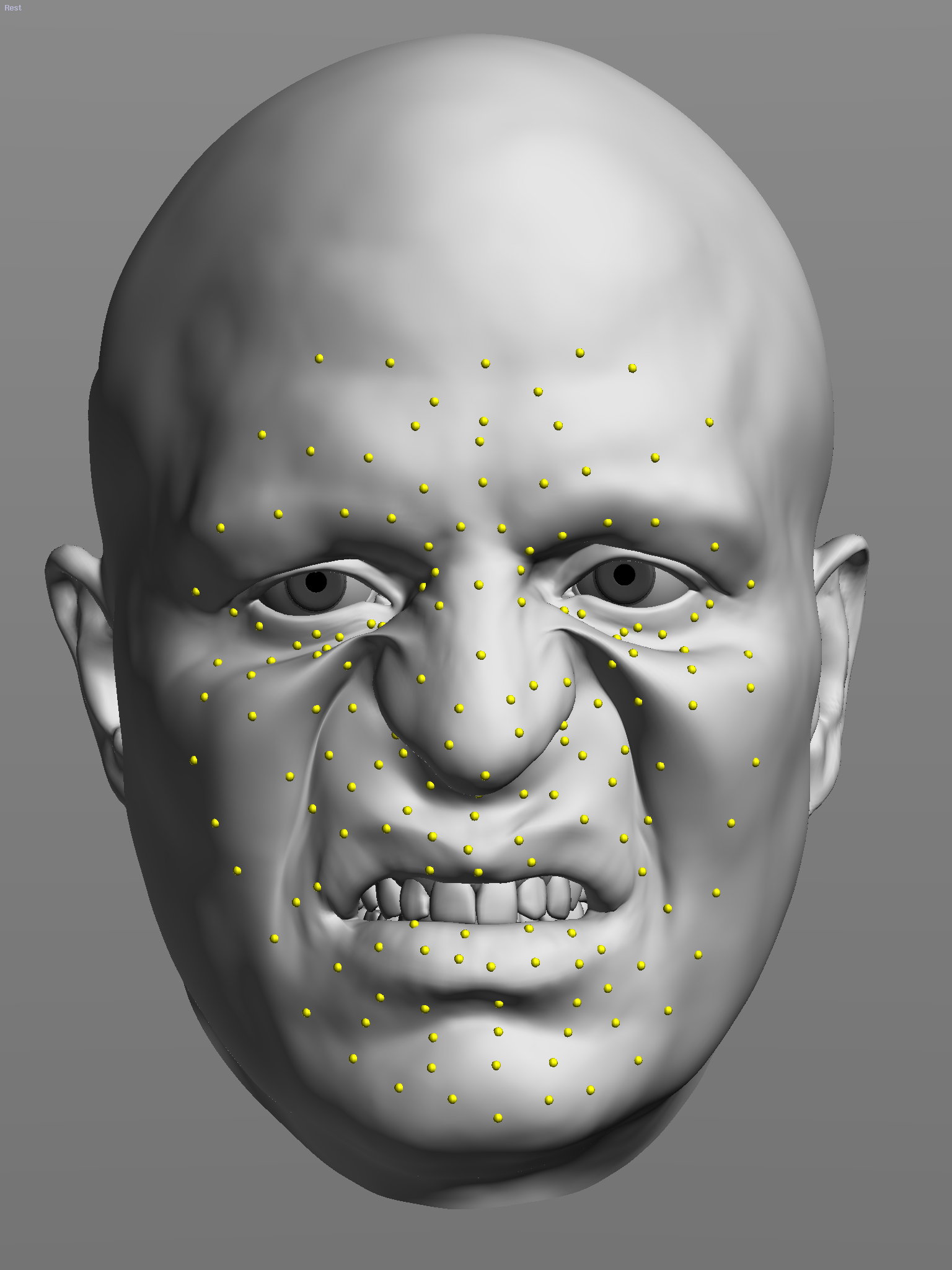}{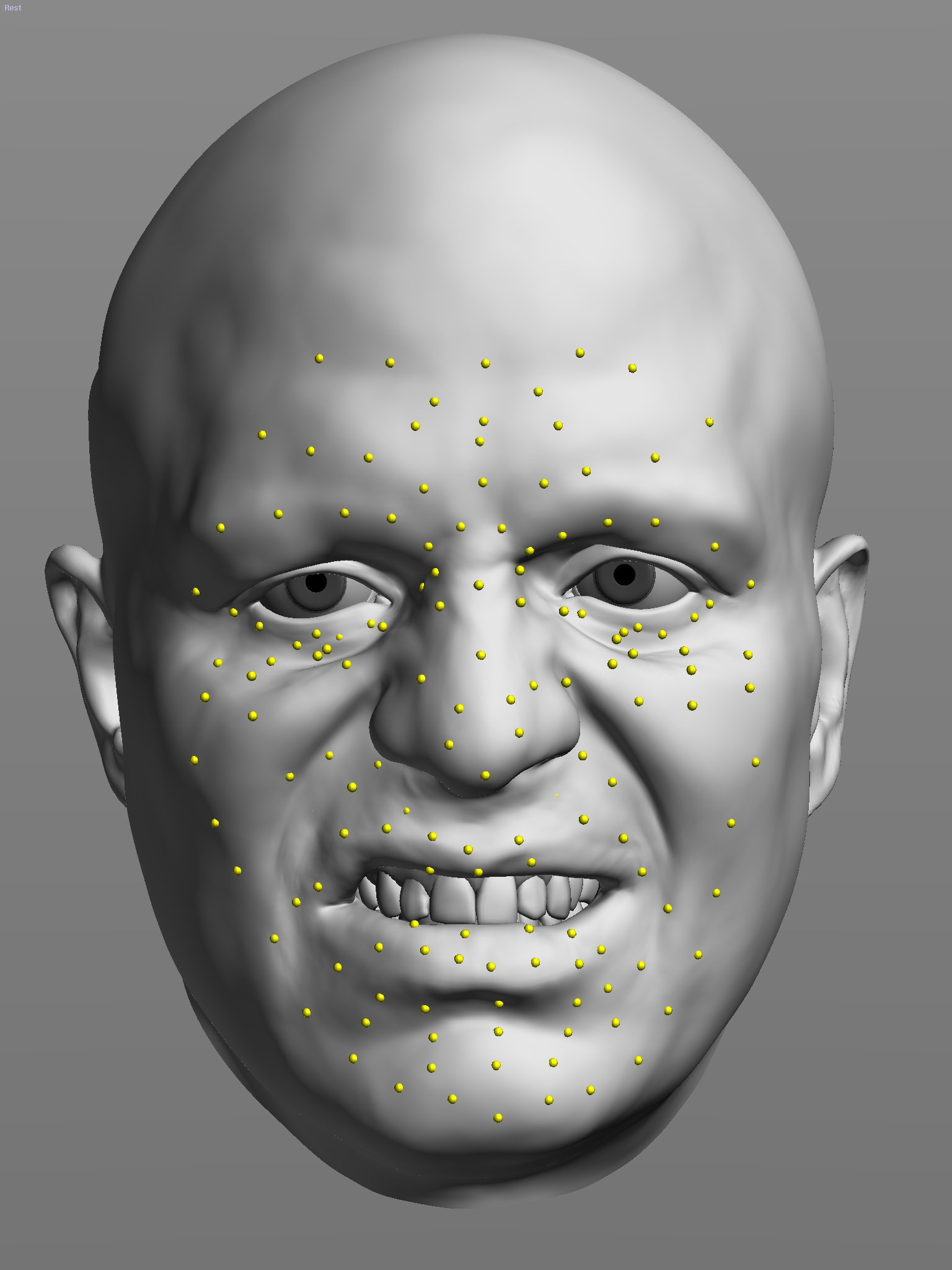}
    \caption{Left: The combination of sneer, snarl, and upper lip raiser blendshapes leads to severe pinching artifacts in the cheeks and excessive deformation in the nose. These blendshapes are often used in conjunction to animate an angry face. Right: Reconstruction obtained using our local geometric indexing algorithm with the bundles calculated from (Left) as input. The reconstruction fixes the aforementioned artifacts in the cheek and nose, improves the shape of the upper lip, and preserves the emotional intent associated with each of the individual blendshapes. \label{fig:correctives}}
\end{figure}

\vspace{-0.1in}
\paragraph{Data Augmentation via Simulation:} 
Figure~\ref{fig:hmc_temporal_smoothness} also illustrates the efficacy of augmenting the dataset using the art-directed muscle simulation framework of \cite{cong2016art}.
Figure~\ref{fig:hmc_temporal_smoothness} (Middle Left) was the result obtained without augmenting and Figure~\ref{fig:hmc_temporal_smoothness} (Middle Right) was the improved result obtained by adding a number of new facial shapes via \cite{cong2016art} as outlined in Section~\ref{sec:dataset}.

\vspace{-0.1in}
\paragraph{Generating/Correcting Rigs:} 
Our local geometric indexing algorithm can also be used to generate actor-specific facial rigs.
Given a generic template blendshape rig applied to the actor neutral mesh, we evaluate bundle positions for individual blendshapes and use these bundle positions as input into our local geometric indexing algorithm to reconstruct corresponding actor-specific blendshapes.
We apply the same approach to combinations of blendshapes in order to obtain corresponding actor-specific corrective shapes \cite{bhat2013high} that do not exhibit the artifacts commonly found in combinations of blendshapes.
See Figure~\ref{fig:correctives}.
These actor-specific blendshapes and corrective shapes can be incorporated into an actor-specific nonlinear blendshape facial rig for use in keyframe animation and other facial capture applications.

\vspace{-0.1in}
\paragraph{Usability:} 
Our local geometric indexing calculations can be performed independently for each bundle and as expected, our parallel CPU implementation using Intel Threading Building Blocks scales linearly.
Given this degree of parallelism in the local geometric indexing scheme as well as the GPU implementation of natural neighbor interpolation demonstrated in \cite{park2006discrete}, our algorithm has the potential to run at interactive rates on the GPU.
Already, our approach is general and efficient enough to have been incorporated for use in the production of a major feature film.
It has been tested on a wide range of production examples by a number of different users with significant creative and technical evaluation on the results.
We show a small selection of our test results in Figure~\ref{fig:hmc_examples}.





\section{Conclusion}

We have presented a data-driven approach for high-resolution facial reconstruction from sparse marker data.
Instead of fitting a parameterized (blendshape) model to the input data or smoothly interpolating a surface displacement to the marker positions, we use a local geometric indexing scheme to identify the most relevant shapes from our dataset for each bundle using a variety of different criteria.
This yields local surface geometry for each bundle that is then combined to obtain a high-resolution facial reconstruction.

We have applied our method to real-world production helmet mounted camera footage to obtain high-quality reconstructions.
Rotoscoped features, including lip occlusion contours, can be readily incorporated as a postprocess.
Finally, our approach has already been deployed for use in a film production pipeline for a major feature film where it has been leveraged by many users to obtain production quality results.




\section*{Acknowledgements}

We would like to thank Cary Phillips, Brian Cantwell, Kevin Sprout, and Industrial Light \& Magic for supporting our research into facial performance capture.


{\small
\bibliographystyle{ieee}
\bibliography{references}
}

\end{document}